\documentclass{article}

\usepackage[preprint]{icml2026}

\usepackage{amsmath}
\usepackage{amssymb}
\usepackage{graphicx}
\usepackage{booktabs}
\usepackage{tabularx}
\usepackage{array}
\usepackage{longtable}             
\usepackage{pdflscape}             
\usepackage{enumitem}
\usepackage{csquotes}
\usepackage{etoolbox}
\usepackage[most,listings]{tcolorbox}
\usepackage{xcolor}
\usepackage[colorlinks=true, allcolors=blue]{hyperref}
\usepackage[nameinlink]{cleveref}
\usepackage{multirow}
\graphicspath{{figures/}}
\usepackage{bm}

\definecolor{inlinecodetext}{HTML}{B42318}
\definecolor{inlinecodebg}{HTML}{F2F4F7}
\renewcommand{\texttt}[1]{%
  {\setlength{\fboxsep}{2pt}%
   \colorbox{inlinecodebg}{\textcolor{inlinecodetext}{\ttfamily #1}}}%
}

\definecolor{codebodybg}{HTML}{F2F4F7}
\definecolor{codetitlebg}{HTML}{EAECF0}
\definecolor{codeframe}{HTML}{D0D5DD}

\tcbset{textbox/.style={
  fontupper=\small\ttfamily,
  fonttitle=\small\ttfamily\bfseries,
  colback=codebodybg,
  colframe=codeframe,
  colbacktitle=codetitlebg,
  coltitle=black,
  boxrule=0.5pt,
  arc=2mm,
  left=6pt,
  right=6pt,
  top=6pt,
  bottom=6pt,
  breakable,
}}

\usepackage{pifont}

\newcommand{\xmark}{\ding{55}}

\usepackage{multibib}
\newcites{evals}{Eval-setup references}
\bibliographystyleevals{unsrtnat}

\AddToShipoutPictureBG*{%
 \AtPageUpperLeft{%
  \hspace{0.78in}%
  \raisebox{-0.65in}{%
   \includegraphics[height=0.35in]{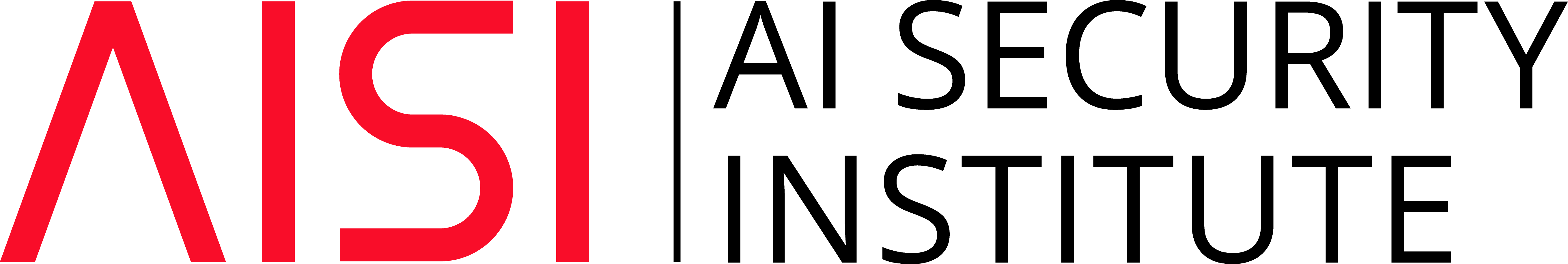}%
  }%
 }%
}

\icmltitlerunning{How Inference Compute Shapes Frontier LLM Evaluation}

\begin{document}

\icmltitle{How Inference Compute Shapes Frontier LLM Evaluation}

\begin{icmlauthorlist}
\icmlauthor{Jessica McFadyen}{aisi}
\icmlauthor{Ole Jorgensen}{aisi,oxford}
\icmlauthor{Harry Coppock}{aisi}
\icmlauthor{Kevin Wei}{aisi,harvard}
\icmlauthor{Cozmin Ududec}{aisi}
\end{icmlauthorlist}

\icmlaffiliation{aisi}{AI Security Institute, London, UK}
\icmlaffiliation{oxford}{University of Oxford, Oxford, UK}
\icmlaffiliation{harvard}{Harvard University, Cambridge, MA, USA}
\icmlcorrespondingauthor{Jessica McFadyen}{jessica.mcfadyen@dsit.gov.uk}

\vskip 0.3in

\printAffiliationsAndNotice{}

\begin{abstract}
AI evaluations are shifting toward harder tasks that benefit from longer trajectories involving tool use and iterative problem solving. As a result, performance is increasingly sensitive to the amount and allocation of compute available at test time (``inference compute''). Yet many evaluations still report performance at a single restrictive budget, meaning that low scores may reflect the evaluation setup rather than the model’s underlying capability. To test this, we evaluate up to 12 frontier language models on seven challenging benchmarks spanning software engineering, mathematics, medicine, and cybersecurity. We use a controlled setup combining three simple inference-scaling interventions: larger token budgets, context compaction, and repeated submission attempts, guided either by the model itself or by minimal correctness feedback. We find three main results. First, larger token budgets substantially improve performance on benchmarks across multiple domains, including cybersecurity, FrontierMath, Humanity’s Last Exam, and TerminalBench. Second, fixed-budget evaluations can increasingly understate frontier capability as models advance. Newer models reach higher performance at large budgets, where they unlock harder tasks and solve them more reliably. Third, benchmarks differ in which inference-scaling methods help most: repeated submission broadly improves performance, but the value of larger token budgets, external feedback, and parallel attempts varies by benchmark. Overall, our results show that benchmark scores are protocol-dependent. We therefore argue that evaluations should report capability as a function of inference-time compute, specify protocol choices explicitly, and compare model generations over a large shared compute range at matched budgets, especially in safety- or policy-relevant settings.
\end{abstract}

\section{Introduction}

As frontier AI benchmarks saturate, evaluations are shifting toward harder, longer-horizon tasks that benefit from extended trajectories, multi-step planning, tool use, and interaction with complex environments \citep{phan2025hle,glazer2024frontiermath,folkerts2026cyber,terminalbench2026,swebenchpro2025,kapoor2026openworld}. Performance on these tasks increasingly depends on how much inference-time compute evaluations allow \citep{ord2025inference,iasr2025}. Yet many evaluations still use modest token budgets, give models only one chance to submit a solution, and report a single protocol-dependent score -- the equivalent of evaluating a human expert under severe time pressure. This risks understating model capability, since a failure may mean the model simply ran out of inference budget rather than that it could not solve the task. 

Here, we systematically study this scaling of performance with inference-time compute (hereafter, ``inference scaling'') within a unified experimental framework. We evaluate six frontier language models spanning multiple generations, released between May 2025 and March 2026, on five challenging benchmarks across software engineering, mathematics, and medicine, and additionally draw on two closely related cybersecurity evaluations from the UK AI Security Institute spanning an overlapping set of 10 models \citep{aisi2026mythos,folkerts2026cyber,aisi2026scaling}. Across these settings, we apply a consistent inference-scaling protocol: expanded total token budgets (one to three orders of magnitude above published benchmark defaults), context compaction, and unlimited submission attempts with or without minimal correctness feedback. The techniques are intentionally straightforward so that we establish a lower bound on what simple, general, and reproducible inference-scaling methods can \emph{elicit}, rather than aiming to maximally elicit each model with benchmark-specific scaffolding.

\begin{figure*}[!h]
  \centering
  \includegraphics[width=\linewidth]{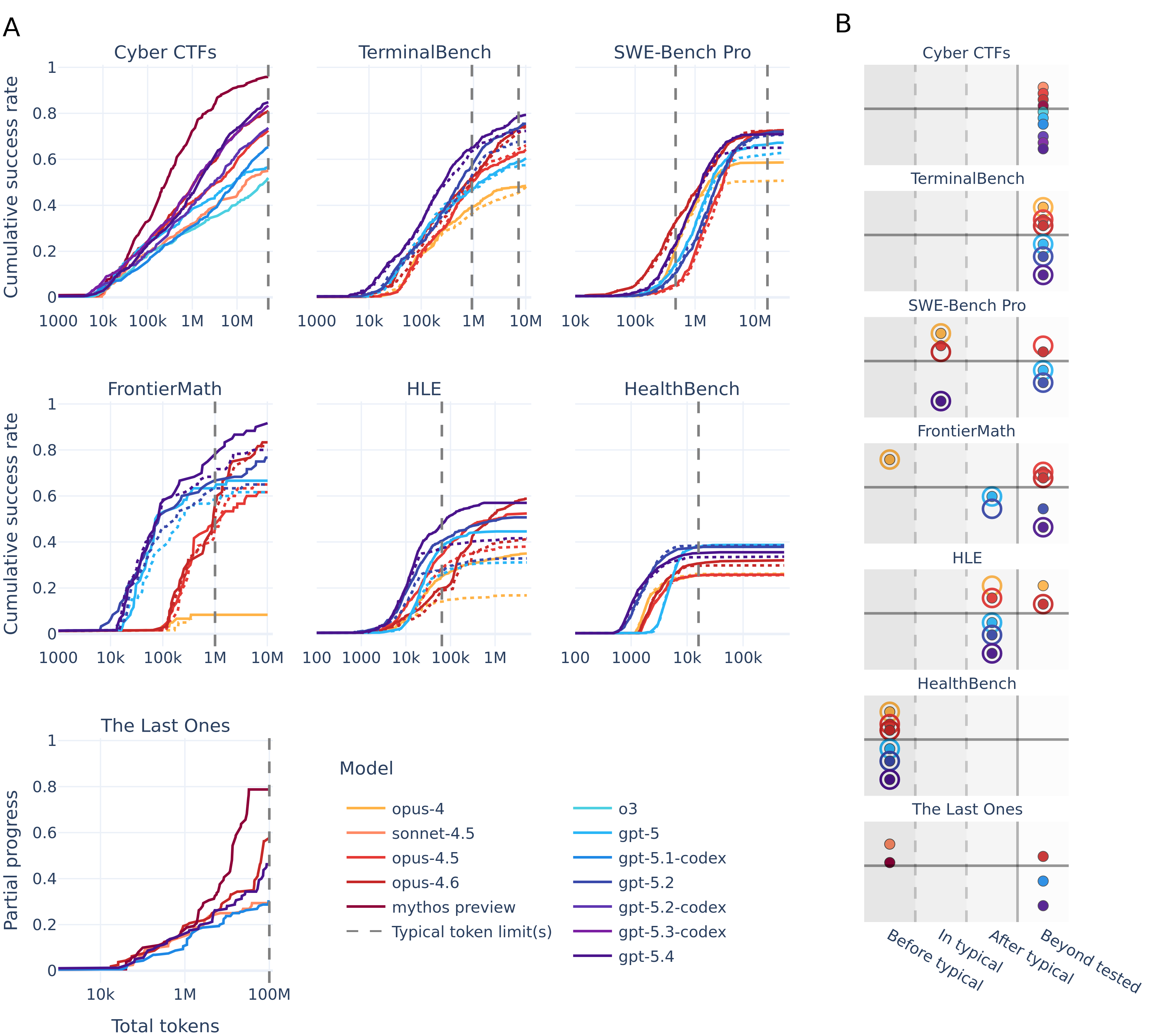}
  \caption{%
   \textbf{Inference scaling curves and plateau locations for frontier models tested on seven benchmarks.}%
   \textbf{(A)} Cumulative performance as a function of total tokens used (excluding LLM judge tokens), with one curve per (model, benchmark, condition) on each benchmark: solid lines for oracle score feedback, dashed for no feedback; each condition includes 5 independent trajectories per task. The two cyber benchmarks use previously collected data under a closely similar oracle-scored protocol only (5 trajectories per task) and are shown with solid lines only. The y-axis reports cumulative mean score: continuous scores in $[0,1]$ for \emph{HealthBench}, partial progress for \emph{The Last Ones} as the proportion of 32 steps completed, and binary 0/1 task success otherwise. The legend orders models by chronological release date within each provider. Vertical dashed lines mark the typical published token budget for each benchmark (two lines indicating a typical range); vertical solid lines mark the token limits imposed in this evaluation.%
   \textbf{(B)} Token budget at which each inference-scaling curve plateaus, bucketed relative to the typical published budget and the tested range: \emph{Before typical}, \emph{In typical} (for benchmarks with a range of typical values), \emph{After typical} (between the typical budget and our tested limit), or \emph{Beyond tested} (no plateau within the tested budget). Points represent models, filled for oracle score feedback and hollow for no feedback on the five main benchmarks; the two cyber benchmarks contribute oracle score feedback points only. Horizontal line separates Anthropic (top) from OpenAI (bottom) models.
  }
  \label{fig:scaling-curves}
\end{figure*}

We find three main patterns:

\begin{enumerate}
    \item \textbf{Inference scaling is substantial but highly benchmark-dependent.} Some benchmarks continue to improve well beyond typical published token budgets -- including FrontierMath, TerminalBench, and Humanity’s Last Exam (HLE) -- while others show weaker marginal gains under our protocol.
    \item \textbf{Newer model generations usually achieve higher performance at large budgets, where they unlock harder tasks and solve them more reliably.} These findings suggest that low-budget evaluations may fail to track progress in the ability to convert additional inference-time compute into performance, and therefore may fail to elicit capabilities that are visible only at larger budgets. Fixed-budget scores therefore give an incomplete picture of the performance frontier reachable under broader inference-time budgets, and this omission may grow as models advance.
    \item \textbf{Gains from inference scaling do not arise from one universal intervention.} Benchmark scores depend partly on protocol choices about if and how models can iterate on their solutions, and whether compute is allocated to a single deep trajectory (“serial” scaling) or spread across multiple shallower ones (“parallel” scaling). Repeated submissions materially improve performance on all benchmarks, and feedback on submission correctness matters most where it can guide continued search (HLE and SWE-Bench Pro). Parallel scaling is strongest on stateless benchmarks (HealthBench and HLE, which do not involve a persistent interactive environment) and weakest on stateful ones. Together, these results demonstrate that different tasks respond to distinct ways of allocating inference-time compute, implying that elicitation is partly protocol-dependent.
\end{enumerate}

Overall, these findings suggest that frontier capability cannot be fully characterized by a single benchmark score measured under a single inference-time protocol. Observed performance depends not only on the model, but also on how much inference-time compute it is given and how that compute is allocated. As a result, evaluations should (i) report capability as a function of inference-time compute rather than as a single fixed-budget number, (ii) treat protocol choices as part of the evaluation design and report them explicitly, and (iii) control for the compute range and protocol when comparing capability across model generations, particularly in safety-critical or policy-relevant contexts \citep{cerruti2026safety_eval_ttc}.

\section*{Background}

Existing evidence suggests that additional inference-time compute can improve performance on difficult evaluations, including cybersecurity \citep{folkerts2026cyber,meta2026musespark}, software engineering \citep{ma2025thinking,ding2026swereplay,epoch2026mirrorcode}, mathematics \citep{muennighoff2025s1,wu2024inference}, medicine \citep{huang2025m1,byun2026}, and other interactive tasks \citep{anthropic2026mythos,wei2025browsecomp}. Inference scaling can be characterised by a cumulative success curve, which shows performance as a function of tokens consumed rather than at one fixed budget. Such curves reveal whether success continues to improve as more inference-time compute is allocated \citep{cerruti2026safety_eval_ttc}, and differences between frontier models often emerge more clearly at high compute levels \citep{folkerts2026cyber}. If performance is still rising across the tested range, the evaluation has measured only part of the achievable performance under that protocol rather than the full limit \citep{folkerts2026cyber,epoch2026mirrorcode}. This is especially important for cross-generation comparisons, since fixed-budget evaluations may miss improvements in how newer models use additional inference-time compute \citep{aisi2026scaling}.

Weak or absent inference scaling can mean either that additional inference compute genuinely does not help in that setting or that the apparent limit is induced by the evaluation protocol. Longer trajectories can hurt performance \citep{gema2025inverse,laban2025lost}, and some domains may be intrinsically less responsive to additional inference compute \citep{sprague2024cot}. But weak scaling can also result from tight budgets, turn caps, timeouts, poor context management, or limited opportunities for iterative refinement \citep{jurkovic2026metr,terminalbench2026,sun2025contextfolding}. These protocol choices restrict both serial depth \textit{within} a trajectory and parallel breadth \textit{across} independent trajectories \citep{wang2023selfconsistency,snell2024scaling}. Evaluations with similar cumulative success curves may in fact permit very different opportunities for search, recovery, or refinement. Moreover, apparent plateaus in these curves may reflect turn caps or timeouts rather than a true limit to productive inference-time compute. Whether an evaluation observes inference scaling or not therefore cannot be interpreted without considering which inference-time opportunities the protocol permits. Assessing what additional inference-time compute enables models to achieve may require a more decomposed analysis of how inference-time compute is allocated and what forms of search, interaction, or refinement the protocol actually enables \citep{marchand2026sandbox}.

Two gaps limit what can be concluded from current evidence. First, existing studies rarely characterise performance over a wide enough inference-compute range. This is costly but necessary to distinguish near-saturation from continued improvement. Both ends of the curve matter: the high-compute tail is informative about what well-resourced actors could achieve, while the low-compute regime is informative about accessibility and less-resourced misuse. Second, evaluators seldom apply a sufficiently comparable setup across benchmarks or model generations within a single framework. This makes it hard to separate differences due to models and tasks from differences due to scaffolding, tools, and protocol. Comparability does not, however, guarantee full elicitation: even under a shared framework, weak scaling may still reflect genuine capability limits or a protocol that fails to elicit capabilities the model could exhibit under a different allocation of inference-time compute.

\section{Methods}

Our \textbf{main experiment} follows a fully crossed design with six frontier models (\Cref{tab:coverage}) evaluated on five non-cyber benchmarks under two feedback conditions, using one shared ReAct-style scaffold \citep{yao2023react} implemented in Inspect AI \citep{inspectai}. For each task in each (benchmark, model, condition) cell, we run 5 independent trajectories. Per-benchmark dataset loading, sandbox configuration, scoring details, and additional protocol specifics are reported in \Cref{app:benchmark-details,app:conditions-prompts,app:repetition-guard}.

We also include two previously collected \textbf{cyber benchmarks} \citep{aisi2026mythos, folkerts2026cyber, aisi2026scaling} in the inference-scaling analyses only. These are not part of the fully crossed main design and cover a different set of models spanning a similar release-date range (\Cref{tab:coverage}). They were run with 5 independent trajectories per task under a near-identical high-budget protocol described below in \Cref{sec:inference-scaling-techniques}.

\subsection{Models}

For the main benchmark suite, we evaluate six frontier models spanning three generations (from May 2025 to March 2026) drawn from two model families (\Cref{tab:coverage}). Within each family, three successive releases (Opus 4 $\to$ 4.5 $\to$ 4.6 and GPT-5 $\to$ 5.2 $\to$ 5.4) enable controlled cross-generation comparisons while holding scaffold, prompts, and inference-time compute budgets fixed. All models run with high reasoning effort (\texttt{xhigh} for Anthropic, \texttt{high} for OpenAI) and a reasoning-token budget of 16{,}000 per generation call -- the budget at which Humanity's Last Exam accuracy is reported to peak \citep{hle2026nature}. This reasoning-token budget is separate from the per-trajectory total budget and governs each individual model call.

Data for the two cyber benchmarks span a different set of models: 10 for the capture-the-flag (CTF) suite (April 2025 to April 2026) and 5 for The Last Ones (September 2025 to April 2026). These both include Mythos Preview, a frontier model checkpoint that Anthropic provided to the UK AI Security Institute (AISI) for evaluation \citep{anthropic2026mythos}. AISI tested two such checkpoints; the one we evaluate is the newer of the two, postdating the checkpoint in AISI's initial Mythos Preview evaluation \citep{aisi2026mythos} and matching the more recent reporting in which this newer checkpoint was the first model to solve both AISI cyber ranges end-to-end \citep{aisi2026doubling}.

\begin{table}[!h]
\centering
\begin{tabular}{l l ccccc | cc}
\toprule
& & \multicolumn{5}{c}{Main benchmark suite} & \multicolumn{2}{c}{Cyber benchmarks} \\
\cmidrule(lr){3-7} \cmidrule(lr){8-9}
Model & Release & TB & SBP & FM & HB & HLE & CTF & TLO \\
\midrule
o3       & Apr, 2025 & \xmark   & \xmark   & \xmark   & \xmark   & \xmark   & \checkmark & \xmark   \\
Opus 4     & May, 2025 & \checkmark & \checkmark & \checkmark & \checkmark & \checkmark & \xmark   & \xmark   \\
GPT-5     & Aug, 2025 & \checkmark & \checkmark & \checkmark & \checkmark & \checkmark & \checkmark & \xmark   \\
Sonnet 4.5   & Sep, 2025 & \xmark   & \xmark   & \xmark   & \xmark   & \xmark   & \checkmark & \checkmark \\
Opus 4.5    & Nov, 2025 & \checkmark & \checkmark & \checkmark & \checkmark & \checkmark & \checkmark & \xmark   \\
GPT-5.1 Codex & Nov, 2025 & \xmark   & \xmark   & \xmark   & \xmark   & \xmark   & \checkmark & \checkmark \\
GPT-5.2    & Dec, 2025 & \checkmark & \checkmark & \checkmark & \checkmark & \checkmark & \xmark   & \xmark   \\
GPT-5.2 Codex & Dec, 2025 & \xmark   & \xmark   & \xmark   & \xmark   & \xmark   & \checkmark & \xmark   \\
Opus 4.6    & Feb, 2026 & \checkmark & \checkmark & \checkmark & \checkmark & \checkmark & \checkmark & \checkmark \\
GPT-5.3 Codex & Feb, 2026 & \xmark   & \xmark   & \xmark   & \xmark   & \xmark   & \checkmark & \xmark   \\
GPT-5.4    & Mar, 2026 & \checkmark & \checkmark & \checkmark & \checkmark & \checkmark & \checkmark & \checkmark \\
Mythos Preview & Apr, 2026 & \xmark   & \xmark   & \xmark   & \xmark   & \xmark   & \checkmark & \checkmark   \\
\bottomrule
\end{tabular}
\caption{Model coverage across the seven benchmarks. Checkmarks indicate the (model, benchmark) cells we evaluated. Models are ordered by release date. The five main benchmarks (TB: TerminalBench; SBP: SWE-Bench Pro; FM: FrontierMath; HB: HealthBench; HLE: Humanity's Last Exam) enter all downstream analyses. The two cyber benchmarks (CTF: Capture the Flag; TLO: The Last Ones) enter the inference-scaling analyses only.}
\label{tab:coverage}
\end{table}

\subsection{Inference-scaling techniques}
\label{sec:inference-scaling-techniques}

We apply three deliberately simple inference-scaling techniques uniformly across every (benchmark, model, condition) cell:
\begin{enumerate}
    \item \textbf{Expanded total token budgets} of 5M--30M tokens per trajectory (\Cref{tab:benchmarks}), sized to task complexity and one to three orders of magnitude above typical benchmark defaults.
    \item \textbf{Context compaction}, which replaces earlier turns with a model-generated summary to enable serial scaling beyond the nominal context-window size. We adopt Inspect AI's default summary-based strategy \citep{inspectai}, triggered when the running context exceeds 130k tokens (about 65\% of the smallest 200k context window across models).
    \item \textbf{Iterative resubmission} with a hard cap of 999 submissions per trajectory, allowing the model either to refine its previous answer or to try a substantially different approach between submissions. To reduce unproductive loops of near-identical submissions, we also employ a lightweight \textbf{repetition guard} that terminates a trajectory when a separate LLM judge finds three or more consecutive submissions semantically equivalent (\Cref{app:repetition-guard}).
\end{enumerate}

These techniques all operate within a single trajectory and target \emph{serial} inference scaling; we study \emph{parallel} inference scaling separately in \Cref{sec:parallel-scaling}, by reanalysing the same trajectories under different fixed-total-budget allocations.

\subsection{Feedback conditions}
\label{sec:feedback-conditions}

Whether a model can productively use a large token budget depends strongly on the feedback available during the trajectory \citep{balachandran2025inferencetime}. We therefore run two conditions for each (model, benchmark) pair:
\begin{enumerate}
    \item \textbf{No feedback}. The model receives only an ambiguous acknowledgement after each submission (\emph{``Your answer has been saved''}) with no indication of correctness, and the trajectory continues until the budget is exhausted or the repetition guard fires. This isolates the model's own ability to judge when it has solved a task.
    \item \textbf{Oracle score feedback}. The model is told whether each submission is correct (or, for HealthBench, given a partial-credit score). The trajectory terminates on the first fully correct submission, as the model is thereafter aware that it has solved the task. 
\end{enumerate}

Both conditions share an adaptive continuation prompt that invites the agent either to refine its previous answer or to try a substantially different approach. The full prompt text and benchmark-specific feedback variants are reproduced in \Cref{app:conditions-prompts}. Each (benchmark, model, task) cell contains 5 trajectories per condition (10 in total).

Unless otherwise stated, token budgets are all-inclusive for the target model and cover input, output, and reasoning tokens; LLM judge tokens are excluded from these budgets and from the token counts shown in scaling plots, but we report when judges are used. A 90-minute timeout is imposed on each individual model generation call. Any trajectories hitting this timeout were re-run, so no trajectories in the final dataset are affected.

The cyber benchmarks were run under a closely matching oracle-feedback protocol, where models were told if they had successfully completed the task or not, without any partial-progress signals.

\subsection{Benchmarks and scoring}
\label{sec:benchmarks-agent-setup}

We select five benchmarks that are both challenging for current frontier models and diverse in domain and task structure (\Cref{tab:benchmarks}). Three are "stateful" benchmarks requiring environment state persistence across multi-turn tool use: \textit{TerminalBench 2.0} \citep{terminalbench2026}, \textit{SWE-Bench Pro} \citep{swebenchpro2025}, and \textit{FrontierMath} \citep{glazer2024frontiermath}. Two are "stateless" knowledge-and-reasoning benchmarks: \textit{HealthBench (Hard)} \citep{openai2025healthbench} and \textit{Humanity's Last Exam (HLE)} \citep{phan2025hle} with multiple-choice questions filtered out. Together they measure terminal-based task execution, real-world software engineering, expert-level mathematics, clinical reasoning, and cross-disciplinary expert knowledge. We additionally reuse data from two cyber benchmarks: \textit{Cyber CTFs} (71 capture-the-flag tasks) and \textit{The Last Ones} (a single long-horizon cyber range scored by 32 milestones toward a final objective). Because these data were not collected under the controlled framework of the main experiment, we use them only for the inference-scaling curves and exclude them from analyses of submission behaviour and feedback. Per-benchmark loading, grading, sandbox configuration, and the cyber data collection protocol are given in \Cref{app:benchmark-details}.

The stateful benchmarks are given \texttt{bash} and \texttt{python} tools, with a custom \texttt{submit\_answer} tool for FrontierMath's code-based submissions. The stateless benchmarks are tool-less and submit via a generic \texttt{submit} tool. Scoring is by programmatic verification for TerminalBench (bundled unit tests), SWE-Bench Pro (bundled test harness), FrontierMath (sandboxed code execution against per-problem reference implementations), Cyber CTFs (flag capture), and The Last Ones (milestone progress), while HLE and HealthBench are graded by a separate LLM judge (GPT-4o-mini at temperature 0). All benchmarks use binary scores, except for HealthBench which is scored continuously against per-conversation physician-designed rubrics. For each benchmark, we use a canonical task set of up to 100 tasks fixed deterministically across all runs. Together with our expanded-budget protocol, this subsampling means absolute scores on the non-cyber benchmarks are not directly comparable to published benchmark results.

\subsection{Typical-budget reference points}
\label{sec:typical-budgets}

To quantify the uplift from typical published budgets to our expanded budgets (\Cref{tab:budget-uplift}), we define a benchmark-specific reference for what counts as a typical stopping point. Published evaluations express limits in different units (output tokens, turn caps, or wall-clock time), which we convert to approximate per-trajectory total-token budgets on a common scale. For HLE, HealthBench, and FrontierMath we use publicly reported token limits (64k, 16k, and 1M total tokens respectively). For TerminalBench and SWE-Bench Pro, which do not report token budgets directly, we estimate token-equivalent stopping points from our own trajectories by measuring cumulative tokens consumed at the point at which the published wall-clock or turn limit would have applied. These converted values are used only as descriptive reference points, to mark typical published budgets in \Cref{fig:scaling-curves}A and to quantify uplift in \Cref{tab:budget-uplift}. Full sources, conversions, and resulting estimates are given in \Cref{app:typical-budgets}.

\begin{table}[htbp]
\centering
\begin{tabularx}{\textwidth}{l >{\raggedright\arraybackslash}X >{\hsize=0.7\hsize\raggedright\arraybackslash}X >{\hsize=1.3\hsize\raggedright\arraybackslash}X r l}
\toprule
Benchmark & Domain & Task type & Scoring & Tasks & Token cap \\
\midrule
TerminalBench & SWE       & Stateful    & Unit tests (binary)    & 86 & 10M \\
SWE-Bench Pro & SWE       & Stateful    & Unit tests (binary)    & 100 & 30M \\
FrontierMath & Mathematics   & Stateful    & Code verification (binary) & 12 & 10M \\
HealthBench  & Medicine     & Stateless QA  & Rubric (continuous)    & 100 & 10M \\
HLE      & Expert knowledge & Stateless QA  & LLM-graded (binary)    & 100 & 5M  \\
Cyber CTFs  & Cyber      & Stateful    & Flag capture (binary)   & 71 & 50M \\
The Last Ones & Cyber      & Stateful    & 32 milestones (continuous) & 1  & 100M \\
\bottomrule
\end{tabularx}
\caption{Benchmark characteristics and specifications. Token cap refers to the per-trajectory total token budget for the target model (input, output, and reasoning tokens), excluding LLM judge tokens.}
\label{tab:benchmarks}
\end{table}

\section{Results}

We organise the results around our three main findings:
\begin{enumerate}
    \item \textbf{\Cref{sec:inference-scaling-results}, inference scaling curves}: Scaling with larger token budgets is substantial but varies sharply across benchmarks, and we characterise where the cumulative curves plateau.
    \item \textbf{\Cref{sec:unlocks}, mechanistic decomposition}: The cross-generational gains visible in these curves come mainly from greater task reach and reliability rather than improved efficiency, which we establish through a task-level decomposition. 
    \item \textbf{\Cref{sec:intervention-dependence}, feedback and iteration}: These gains do not arise from any single intervention: benchmarks respond differently to repeated submission with feedback and to serial versus parallel allocation of a fixed compute budget.
\end{enumerate}

Throughout the results that follow, weak scaling on a given benchmark should be read as a property of our protocol -- it does not rule out larger gains under different scaffolds, tools, or compute allocations.

\subsection{Inference scaling is substantial but varies across benchmarks}
\label{sec:inference-scaling-results}

To examine performance as a function of total tokens consumed (excluding LLM judge tokens), we compute inference scaling curves as follows. For trajectory $i$, let $s_i$ denote its final credited score -- 1 if any submission is correct and 0 otherwise for the binary benchmarks, and the maximum submission score for HealthBench\footnote{In a secondary analysis of the no-feedback condition, scoring
trajectories by their first or last submission -- arguably more
representative of real-world use -- lowered solve rates on most benchmarks,
most strongly on HealthBench (where models reach a correct answer and then
regress) and not at all on SWE-Bench Pro (where models converge to and stay
at correct answers). The serial and parallel scaling results below were
unchanged under all three scoring rules. Moreover, a small pilot suggests an LLM
best-of-n selector can recover much of the gap on some benchmarks.} -- and let $\kappa_i$ be the total token count at which trajectory $i$ first attains $s_i$. The trajectory inference curve $S_i(t)$ is zero until $s_i$ is first attained and equal to $s_i$ thereafter:
\begin{equation}
S_i(t) := s_i \,\mathbf{1}\!\left(\kappa_i \le t\right),
\label{eq:trajectory-equation}
\end{equation}
with $S_i(t)\equiv 0$ when $s_i=0$. The aggregate inference curve is the mean of $S_i(t)$ over a specified set of trajectories,
\begin{equation}
S_{\mathrm{agg}}(t) := N^{-1}\sum_i S_i(t).
\label{eq:aggregate-inference-equation}
\end{equation}

\Cref{fig:scaling-curves}A plots the resultant inference scaling curves, with separate curves per no feedback and oracle score feedback conditions. The quantitative summaries in this subsection pool trajectories across both feedback conditions and across tasks, characterising overall performance under our full evaluation protocol rather than any single feedback setting.

\subsubsection{Typical budgets can miss substantial additional gains}

\paragraph{Software engineering benchmarks show little headroom beyond typical budgets.}
For the two software engineering benchmarks, which are already commonly evaluated at relatively large budgets, extending further yields only limited improvement (\Cref{tab:budget-uplift}). SWE-Bench Pro is typically evaluated at up to about 16M total tokens (equivalent to 250 turns). Near-doubling this budget to our 30M cap adds only $+0.3 \pm 0.3$ percentage points across models. TerminalBench is typically evaluated at up to about 7.3M total tokens (3.3-hour limit) and extending to 10M adds only $+1.3 \pm 1.0$ percentage points. This pattern indicates that, for these benchmarks, the typical published budget ranges are already fairly large and that extending them further buys only modest additional performance at substantial per-run compute cost (an extra 14M tokens for SWE-Bench Pro and 2M for TerminalBench). This should be interpreted as limited responsiveness to the specific serial scaling interventions tested here, not as evidence that alternative allocations of inference compute would necessarily yield similarly small gains.

\paragraph{FrontierMath and HLE show substantial headroom beyond typical budgets.}
In contrast, some of the benchmarks with smaller typical budgets see a larger effect from additional compute. FrontierMath's commonly used budget of 1M total tokens \citep{burnham2025frontiermath} is itself an order-of-magnitude increase over the earlier 100k-token scaffold \citep{glazer2024frontiermath}, and extending from 1M to 10M adds a further $+11.7 \pm 11.0$ percentage points on average. HLE also continues to improve well beyond its typical 64k-output-token budget, gaining $+11.9 \pm 10.4$ points up to our 5M cap. This uplift is larger under oracle feedback ($+15.5 \pm 12.3$ points) than under no feedback ($+8.3 \pm 7.5$), consistent with the stronger submission-feedback effects on HLE reported in \Cref{sec:serial-scaling}.

\paragraph{HealthBench shows minimal headroom despite a small typical budget.}
The exception is HealthBench, which shows little change over the same comparison: only $+0.3 \pm 0.4$ points between a typical 16k-output-token budget and our much larger 10M cap. Under our scaffold and grading set-up, this benchmark therefore shows limited sensitivity to the serial inference-scaling interventions tested here within the observed range. This does not rule out larger gains under different scaffolds, stopping rules, or width--depth allocations.

\begin{table*}[!h]
\centering
\begin{tabular}{lcccl}
\toprule
Benchmark & Typical published budget & Evaluated budget & Condition(s) & Mean performance gain \\
\midrule
\multirow{3}{*}{TerminalBench}
  & \multirow{3}{*}{932K--7.3M total tokens} & \multirow{3}{*}{10M total tokens} & \textbf{Pooled}          & $\bm{+1.26 \pm 0.99}$\% \\
  &                                          &                                    & Oracle feedback          & $+1.20 \pm 0.54$\% \\
  &                                          &                                    & No feedback              & $+1.32 \pm 1.36$\% \\
\midrule
\multirow{3}{*}{SWE-Bench Pro}
  & \multirow{3}{*}{473K--16M total tokens}  & \multirow{3}{*}{30M total tokens} & \textbf{Pooled}          & $\bm{+0.27 \pm 0.31}$\% \\
  &                                          &                                    & Oracle feedback          & $+0.27 \pm 0.30$\% \\
  &                                          &                                    & No feedback              & $+0.27 \pm 0.35$\% \\
\midrule
\multirow{3}{*}{FrontierMath}
  & \multirow{3}{*}{1M total tokens}         & \multirow{3}{*}{10M total tokens} & \textbf{Pooled}          & $\bm{+11.67 \pm 10.99}$\% \\
  &                                          &                                    & Oracle feedback          & $+11.94 \pm 10.08$\% \\
  &                                          &                                    & No feedback              & $+11.39 \pm 12.80$\% \\
\midrule
\multirow{3}{*}{HLE}
  & \multirow{3}{*}{64K total tokens}        & \multirow{3}{*}{5M total tokens}  & \textbf{Pooled}          & $\bm{+11.87 \pm 10.40}$\% \\
  &                                          &                                    & Oracle feedback          & $+15.47 \pm 12.26$\% \\
  &                                          &                                    & No feedback              & $+8.27 \pm 7.51$\% \\
\midrule
\multirow{3}{*}{HealthBench}
  & \multirow{3}{*}{16K total tokens}        & \multirow{3}{*}{10M total tokens} & \textbf{Pooled}          & $\bm{+0.32 \pm 0.42}$ pts \\
  &                                          &                                    & Oracle feedback          & $+0.54 \pm 0.51$ pts \\
  &                                          &                                    & No feedback              & $+0.11 \pm 0.17$ pts \\
\bottomrule
\end{tabular}
\caption{Performance gain from each benchmark's typical published budget to the evaluated budget used in this study, shown separately for pooled trajectories, oracle score feedback trajectories, and no feedback trajectories. Gains are averaged across models. Where the published budget spans a range (TerminalBench, SWE-Bench Pro), we report both the lower and upper bounds and compute the gain from the upper bound. Values are reported in percentage points, except for HealthBench, where the native benchmark point scale is used.}
\label{tab:budget-uplift}
\end{table*}

Taken together, these comparisons show that some typical published budgets already capture most of the gains visible under our protocol, whereas others leave substantial performance unrealised.

\subsubsection{Evidence of plateauing differs across benchmarks}
\label{sec:plateau-results}

The cumulative curves in \Cref{fig:scaling-curves}A suggest that some benchmark--model--condition curves begin to plateau within the tested range, while others continue to improve at the largest budget we evaluate. Throughout, we use ``diminishing returns'' descriptively to mean a local plateau or weak remaining growth at the right edge of the tested range, on a log-$x$ scale. To summarise this pattern, we estimate the local rate of improvement at the cap as
\begin{equation}
g_{\mathrm{cap}} \;=\; \frac{y(\mathrm{cap}) - y(\mathrm{cap}/2)}{\log_{10}(2)},
\label{eq:plateau-equation}
\end{equation}
expressed in percentage points per $10\times$ increase in total tokens. As a descriptive convention, ``weak remaining growth'' means \(g_{\mathrm{cap}} < 1\) percentage point per $10\times$ increase. Under this definition, we observe three broad patterns (\Cref{fig:scaling-curves}B).

\paragraph{TerminalBench and cyber benchmarks show continued growth across all or most tested models.}
Three benchmarks continue to rise for most tested models within the observed range: TerminalBench (all six models above threshold at our 10M cap), Cyber CTFs (all ten models), and The Last Ones (three of five models). TerminalBench and the Cyber CTFs are the clearest cases where the tested range still appears insufficient to reveal compute-saturated performance. On The Last Ones, the oldest and newest models -- Sonnet 4.5 and Mythos Preview -- plateau within the tested range, likely for opposite reasons: Sonnet 4.5 appears to hit a capability floor, whereas Mythos Preview plateaus at high (but not maximal) average milestone completion.

\paragraph{FrontierMath, HLE, and SWE-Bench Pro show mixed evidence of plateauing.}
Three benchmarks show mixed signs of plateauing, with different patterns across models. FrontierMath shows the clearest generational pattern, where the two oldest models (Opus 4, GPT-5) plateau earlier, while later models generally continue improving to larger budgets. This split is somewhat weaker under the no feedback condition (\Cref{fig:scaling-curves}B). HLE plateaus consistently late relative to its typical budget: no model plateaus before the typical 64k budget in either condition. Under no feedback, five of the six models plateau between the typical budget and the tested cap, with only Opus 4.6 still improving at the cap. Oracle score feedback shifts plateau locations later for the Opus family: Opus 4 and Opus 4.6 remain beyond the tested range, while the three GPT-family models and Opus 4.5 plateau after the typical budget but within the tested range. SWE-Bench Pro has the least structured pattern: GPT-5 and GPT-5.2 are the only Open AI models above threshold in both feedback conditions (though visual inspection suggests their curves likely flatten shortly beyond the tested range), while Anthropic's Opus 4.5 is above threshold only under no feedback and Opus 4.6 only under oracle feedback (\Cref{fig:scaling-curves}A).

\paragraph{HealthBench shows near diminishing returns across all tested models.}
HealthBench is the only benchmark on which all six tested models plateau within the typical token budget of 16k total tokens. This indicates that, under our scaffold and grading set-up, this benchmark appears close to diminishing returns under the tested protocol, well within a typical evaluation budget.

Overall, these findings indicate that diminishing returns from additional inference-time compute vary substantially across benchmarks, even under a controlled, shared inference-scaling protocol. Whether these inference scaling curves plateau within the tested range can also depend on the model family, model generation, and feedback condition. Additional curve-level summaries reported in \Cref{app:curve-characteristics} show the same broad pattern: newer models achieve higher performance, begin to succeed on tasks at lower token budgets, and -- except on HealthBench -- gain performance more rapidly after onset.

\subsection{Generational gains are driven mainly by reach and reliability rather than efficiency}
\label{sec:unlocks}

The aggregate curve-level patterns above do not reveal whether later generations solve \emph{new} tasks, solve the \emph{same} tasks with fewer tokens, or simply solve already-solvable tasks more consistently across repeated trajectories. To separate these possibilities, we decompose improvement at the task level into three components:
\begin{enumerate}
    \item \textbf{Reach} -- the proportion of tasks a model reliably \textit{unlocks}, meaning that the task is solved at some point within at least two independent trajectories (ruling out lucky one-off successes)
    \item \textbf{Efficiency} -- how many output tokens are needed to solve unlocked tasks
    \item \textbf{Reliability} -- the proportion of trajectories per unlocked task in which a solution is found.
\end{enumerate}

The analysis covers the five main benchmarks plus Cyber CTFs; The Last Ones is excluded because it contains only a single long-horizon task. Task difficulty is defined as the task's median solve rate across models (higher = easier). For the binary benchmarks, a trajectory solves a task if at least one submission receives a score of 1; for HealthBench, submission scores are first binarised using a global median split at 0.38 (the median score achieved across all trajectories). Each component is quantified per benchmark by regressing its outcome -- the unlock indicator for reach (a linear probability model), log tokens-to-solve for efficiency, and solve rate for reliability -- on model generation, task difficulty, and their interaction. Full model specifications are in \Cref{app:task-level-analysis}.

Efficiency and reliability are evaluated only on tasks that a given model unlocks, rather than on a single task set shared by all models. These analyses therefore describe performance conditional on reach: they capture how models behave on tasks they can solve at least once. Because newer models often unlock harder tasks than earlier models, and harder tasks may require more tokens, comparisons of efficiency across generations may understate true gains in token efficiency. To address this, we conduct a sensitivity analysis restricted to a balanced panel of tasks unlocked by all models and find the qualitative pattern is unchanged, indicating that this compositional effect does not drive the main result (\Cref{app:efficiency-balanced-panel}).

\begin{figure*}[!h]
 \centering
 \includegraphics[width=\linewidth]{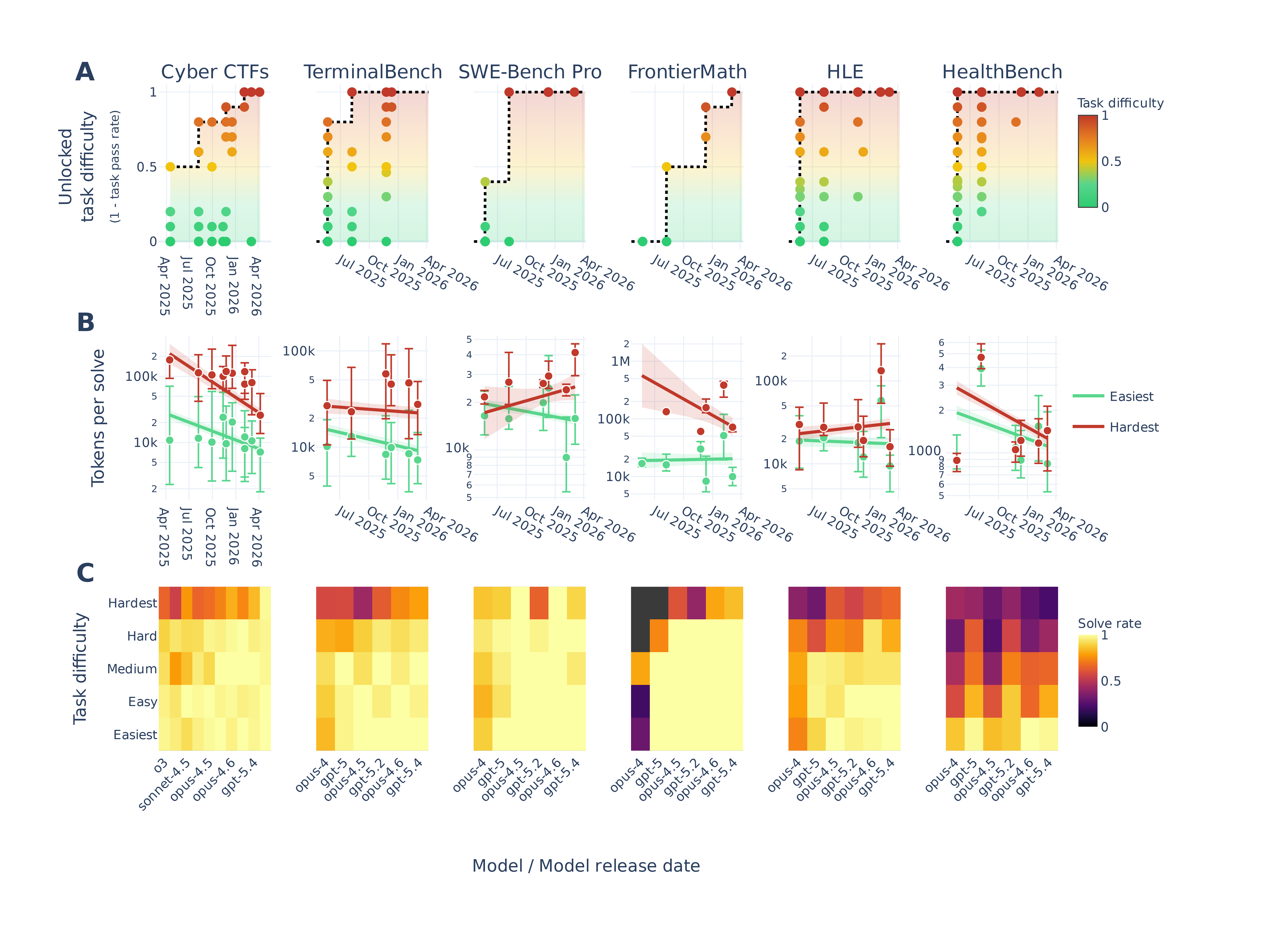}
 \caption{\textbf{Task-level decomposition of generational gains.} \textbf{(A) Reach increases as newer models unlock harder tasks.} Per-task scatter. $x$-axis: release date of the earliest model to unlock the task (solving it in at least 2 of 5 attempts). $y$-axis: per-task difficulty (1 minus the cross-model median solve rate of task $t$; higher = harder). The shaded region beneath the frontier is filled with a difficulty-scaled colour gradient, green for easy and red for hard, reflecting the space of task difficulties unlocked as of each release date. The dotted step line traces the running maximum task difficulty unlocked over time -- the hardest task any model has solved as of each release date. \textbf{(B) Efficiency gains are benchmark-dependent.} Output tokens per solve over model release date, restricted to unlocked (model, task) cells. Lines are simple-slope predictions from a mixed-effects regression fit on continuous task difficulty (see \Cref{app:task-level-analysis}), evaluated at the within-benchmark median split used for visualisation (easiest half in green, hardest half in red); they are therefore not direct fits to the plotted binned medians. Shaded bands: SEM. Dots: observed (model, bin) medians of tokens-to-solve over unlocked tasks, with error bars at the 25th and 75th percentiles. \textbf{(C) Reliability gains are more widespread.} Per-benchmark heat map of within-model mean solve rate over unlocked tasks. Rows: five within-benchmark task-difficulty-quantile bins (hardest at top, easiest at bottom). Columns: models in release order. Cells where the model has no unlocked tasks in that difficulty bin are rendered in grey.}
 \label{fig:task-level-decomposition}
\end{figure*}

\begin{table*}[!h]
\centering
\begin{tabular}{lcccccc}
\toprule
& \multicolumn{2}{c}{Reach} & \multicolumn{2}{c}{Efficiency} & \multicolumn{2}{c}{Reliability} \\
\cmidrule(lr){2-3} \cmidrule(lr){4-5} \cmidrule(lr){6-7}
Benchmark & $\beta_{\mathrm{gen}}$ & $\beta_{\mathrm{int}}$ & $\beta_{\mathrm{gen}}$ & $\beta_{\mathrm{int}}$ & $\beta_{\mathrm{gen}}$ & $\beta_{\mathrm{int}}$ \\
\midrule
Cyber CTFs  & $+0.488^{***}$ & $-0.760^{***}$ & $-0.599^{***}$ & $+0.766^{*}$  & $+0.218^{***}$ & $-0.977^{***}$ \\
TerminalBench & $+0.248^{***}$ & $-0.383^{*}$  & $-0.174^{***}$ & $-0.352$    & $+0.146^{***}$ & $-0.435$ \\
SWE-Bench Pro & $+0.035^{**}$  & $-0.042$      & $-0.092^{**}$ & $-0.432$    & $+0.102^{***}$ & $-0.138$ \\
FrontierMath & $+0.631^{***}$ & $-0.779^{*}$  & $-0.165$    & $+1.323$    & $+0.605^{***}$ & $-1.424^{**}$ \\
HLE      & $+0.188^{***}$ & $+0.000$      & $+0.030$    & $-0.358$    & $+0.223^{***}$ & $-0.194$ \\
HealthBench  & $+0.097^{*}$   & $+0.051$      & $-0.294^{***}$ & $+0.203$    & $+0.009$    & $+0.199^{*}$ \\
\bottomrule
\end{tabular}
\caption{Summary of task-level generational effects. For each component, $\beta_{\mathrm{gen}}$ is the main effect of model generation and $\beta_{\mathrm{int}}$ its interaction with task difficulty, estimated per benchmark; full specifications are given in \Cref{app:task-level-analysis}. \textbf{Reach} coefficients come from a linear probability model for whether a model unlocks a task, so positive $\beta_{\mathrm{gen}}$ indicates that later models unlock more tasks. For \textbf{efficiency}, negative $\beta_{\mathrm{gen}}$ indicates that later models use fewer output tokens to solve unlocked tasks. For \textbf{reliability}, positive $\beta_{\mathrm{gen}}$ indicates that later models solve unlocked tasks more consistently across trajectories. Higher difficulty values denote easier tasks, so negative $\beta_{\mathrm{int}}$ in the reach and reliability analyses indicates gains concentrated on harder tasks. $^{*}p<0.05$, $^{**}p<0.01$, $^{***}p<0.001$}
\label{tab:task-level-summary}
\vspace{2mm}
\end{table*}

\subsubsection{Reach increases with model generation, often more so on harder tasks}

For each benchmark, later model generations unlock a larger proportion of tasks. This is indicated by a positive, significant main effect of model generation on the probability of unlocking a task on all six benchmarks (\Cref{tab:task-level-summary}), with the largest effect on FrontierMath and Cyber CTFs. At the task level, the tasks first unlocked by newer models also tend to be harder (\Cref{fig:task-level-decomposition}A). The generation-by-difficulty interaction is negative -- indicating that reach gains are concentrated on harder tasks -- on four of six benchmarks, and significantly so on Cyber CTFs, FrontierMath, and TerminalBench. HealthBench and HLE are the exceptions, showing no such concentration.

\subsubsection{Efficiency gains are uneven across benchmarks and conditional on reach}

Newer models use fewer output tokens to solve unlocked tasks on four of six benchmarks, indicated by a negative, significant main effect of model generation that is largest on Cyber CTFs and HealthBench and smaller but still significant on TerminalBench and SWE-Bench Pro (\Cref{tab:task-level-summary}; \Cref{fig:task-level-decomposition}B). FrontierMath and HLE show no significant generation effect on efficiency. Collectively, these results indicate that token-efficiency improvements are present but uneven across benchmarks.

Because efficiency is estimated on unlocked \emph{(model, task)} cells, it should be interpreted as conditional on reach rather than on a fixed common task set. Re-fitting on only those tasks unlocked by all models gives similar generation effects on Cyber CTFs, TerminalBench, SWE-Bench Pro, and HealthBench, but less stable generation-by-difficulty interactions (\Cref{app:task-level-analysis}). Generational improvement therefore appears to expand the set of solvable tasks more consistently than it reduces the token cost of solving tasks already within reach.

\subsubsection{Reliability gains are more widespread}

Reliability improves more consistently than efficiency, with newer models solving unlocked tasks more often across repeated trajectories on all benchmarks except HealthBench (\Cref{tab:task-level-summary}; \Cref{fig:task-level-decomposition}C). On Cyber CTFs, FrontierMath, and TerminalBench these gains are concentrated more strongly on harder tasks. HLE shows gains more uniform across difficulty, while SWE-Bench Pro shows a weak negative interaction in the same direction but without clear difficulty-specific concentration. HealthBench is the main exception, with no clear overall reliability gain. A significant interaction between model generation and task difficulty indicates that what little reliability changes with generation does so on easier rather than harder tasks. 

In sum, these results suggest that newer generations improve not only reach but also reliability on reachable tasks -- though whether these gains concentrate on harder tasks depends on the benchmark.

\subsection{Protocol choices shape inference-scaling gains}
\label{sec:intervention-dependence}

The results above show that inference-scaling gains vary across benchmarks and are driven mainly by greater reach and reliability. We therefore examine two protocol dimensions that may shape those gains within our setup: repeated submission under different feedback conditions (\Cref{sec:serial-scaling}), and whether a fixed total budget is concentrated in one deep trajectory or spread across several shallower ones (\Cref{sec:parallel-scaling}).

\begin{figure*}[!h]
 \centering
 \includegraphics[width=\linewidth]{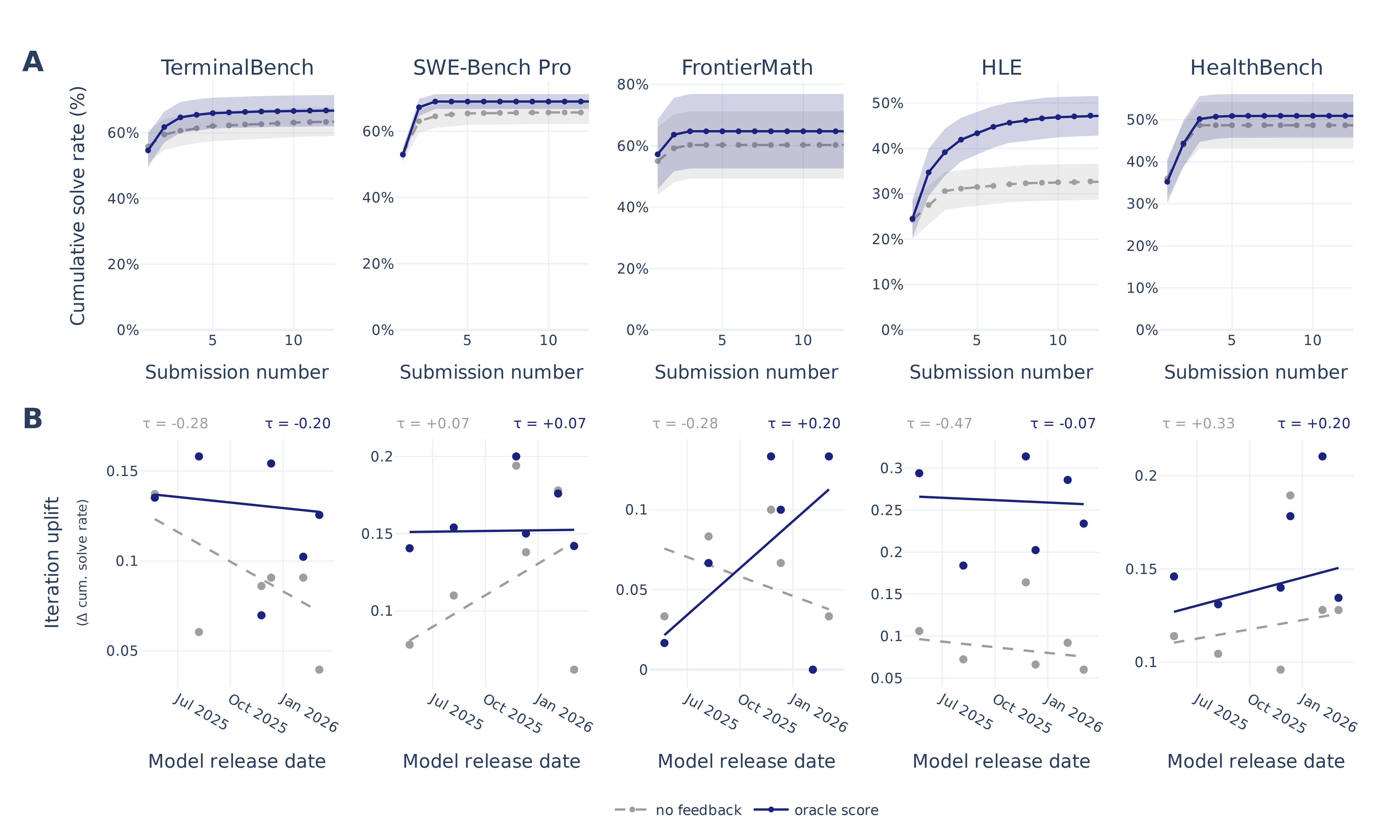}
 \caption{%
  \textbf{Serial scaling under no feedback versus oracle score feedback.}
  (\textbf{A}) Cumulative solve rate versus submission index $k$, pooled with equal weight across models within each benchmark--condition cell. Bands: $\pm 1$ SEM across models. Grey dashed lines: no feedback; navy solid lines: oracle score feedback.
  (\textbf{B}) Iteration uplift $\Delta$ against model release date. Points are models; lines are per-condition Theil--Sen fits (\Cref{app:uplift-analysis}). Per-facet Kendall $\tau$ between release date and $\Delta$ is annotated at the top of each panel (grey: no feedback; navy: oracle score feedback).
 }
 \label{fig:serial-cumulative}
\end{figure*}

\subsubsection{Repeated submission helps most when feedback enables continued search}
\label{sec:serial-scaling}

On longer-horizon tasks, the return to additional serial inference compute may depend on whether models are allowed to resubmit answers within a trajectory and on what feedback they receive after doing so. To test this, we compare the two conditions defined in Methods: no feedback (ambiguous acknowledgement) and oracle score feedback (simple correctness signal). The two cyber benchmarks are excluded because they were collected under only one feedback condition. For comparability across benchmarks, these analyses use binary trajectory outcomes; for HealthBench, which is natively continuously scored, we take the highest-scored submission within each trajectory and binarise using a median split (median score = 0.38).

\paragraph{Repeated submissions improve performance on all benchmarks.}
Allowing repeated submissions improves cumulative performance on all five evaluated benchmarks (\Cref{fig:serial-cumulative}A; \Cref{tab:serial-iteration-main}). Averaged over conditions, cumulative performance rises by $+6.4$ points on FrontierMath, $+10.4$ on TerminalBench, $+14.2$ on HealthBench, $+14.4$ on SWE-Bench Pro, and $+17.3$ on HLE from the first submission to the highest cumulative level reached, corresponding to uplift multipliers from $1.11\times$ on FrontierMath to $1.71\times$ on HLE.

\begin{table}[!h]
\centering
\begin{tabular}{lcccccc}
\toprule
& & \multicolumn{3}{c}{Uplift multiplier} & \multicolumn{2}{c}{$k_{90}$} \\
\cmidrule(lr){3-5} \cmidrule(lr){6-7}
Benchmark & Iteration gain & Overall & No feedback & Oracle feedback & No feedback & Oracle feedback \\
\midrule
FrontierMath  & $+6.39\%$  & $1.11\times$ & $1.10\times$ & $1.13\times$ & 3  & 3 \\
HLE           & $+17.29\%$ & $1.71\times$ & $1.39\times$ & $2.03\times$ & 12 & 13 \\
HealthBench   & $+14.17\%$ & $1.40\times$ & $1.35\times$ & $1.45\times$ & 3  & 3 \\
SWE-Bench Pro & $+14.36\%$ & $1.27\times$ & $1.24\times$ & $1.30\times$ & 4  & 3 \\
TerminalBench & $+10.42\%$ & $1.19\times$ & $1.15\times$ & $1.23\times$ & 14 & 5 \\
\bottomrule
\end{tabular}
\caption{Descriptive summaries of repeated-submission gains. Iteration gain is the increase in cumulative performance, in percentage points, from the first submission to the highest observed cumulative level. Uplift multipliers are reported for the overall, no-feedback, and oracle-score-feedback settings. $k_{90}$ is the number of submissions required to realise 90\% of this gain.}
\label{tab:serial-iteration-main}
\end{table}

\paragraph{Oracle feedback increases iteration gains most where it supports continued search.}
A regression analysis comparing no-feedback and oracle-feedback trajectories within tasks (\Cref{app:serial-regressions}) shows that oracle feedback significantly improves eventual success on HLE, with iteration adding $+25.2$ points under oracle feedback versus $+9.3$ under no feedback ($2.71\times$). Oracle feedback also significantly improves eventual success on SWE-Bench Pro, though with a smaller descriptive uplift ($+16.0$ vs.\ $+12.7$; $1.26\times$). By contrast, FrontierMath, TerminalBench, and HealthBench show no significant benchmark-level success effect of oracle feedback, despite modest descriptive uplift under oracle feedback on FrontierMath ($+7.5$ vs.\ $+5.3$; $1.42\times$), TerminalBench ($+12.4$ vs.\ $+8.4$; $1.48\times$), and HealthBench ($+15.7$ vs.\ $+12.7$; $1.24\times$). On HealthBench, this repeated-submission gain coexists with the weak inference scaling over total tokens consumed reported in \Cref{sec:inference-scaling-results}, suggesting that, under our protocol, iterative refinement can help even when longer token trajectories do not.

\paragraph{Some benchmarks benefit from only a few extra attempts, while others support longer serial search.}
The shape of these gains differs across benchmarks. FrontierMath and HealthBench realise 90\% of their total iteration gain within three submissions in both conditions, and SWE-Bench Pro within three to four, suggesting that repeated submission is useful but shallow in these settings. Consistent with this, most unsolved trajectories on these benchmarks end after the model begins repeating semantically similar answers rather than exhausting the token budget (92--100\% of unsolved trajectories across these three benchmarks; \Cref{tab:serial-exhaustion}), indicating that models typically converge after a small number of distinct attempts rather than sustaining long productive search. By contrast, HLE continues improving for 12--13 submissions, indicating a longer search process composed largely of many short answer attempts. TerminalBench also benefits from more extended search and shows the strongest dependence on feedback, reaching 90\% of its total iteration gain by submission 5 under oracle feedback but requiring up to 14 submissions under no feedback. This matches its much higher rate of budget exhaustion (34\% under no feedback and 39\% of non-correct trajectories under oracle feedback; \Cref{tab:serial-exhaustion}), consistent with longer, more interaction-heavy attempts in which correctness feedback helps models decide sooner when to stop, revise, or switch approaches. Consistent with this interpretation, among trajectories that eventually succeed, oracle feedback does not reliably reduce submissions to the first correct answer, and on HLE and SWE-Bench Pro successful oracle-feedback trajectories use slightly more submissions on average (\Cref{app:serial-regressions}), suggesting that feedback can enable productive continued search rather than merely accelerating convergence.

\begin{table*}[!h]
\centering
\begin{tabular}{llccc}
\toprule
Benchmark & Condition & Correct & Repeated answer & Token budget exhausted \\
\midrule
FrontierMath  & No feedback     & -      & 94.2\% & 5.8\% \\
              & Oracle feedback & 64.7\% & 30.3\% & 5.0\% \\
\midrule
HealthBench   & No feedback     & -      & 100.0\% & 0.0\% \\
              & Oracle feedback & -*    & 99.5\% & 0.5\% \\
\midrule
HLE           & No feedback     & -      & 98.1\% & 1.9\% \\
              & Oracle feedback & 49.5\% & 46.2\% & 4.2\% \\
\midrule
SWE-Bench Pro & No feedback     & -      & 92.1\% & 7.9\% \\
              & Oracle feedback & 68.7\% & 31.0\% & 0.3\% \\
\midrule
TerminalBench & No feedback     & -      & 65.5\% & 34.5\% \\
              & Oracle feedback & 65.8\% & 21.0\% & 13.2\% \\
\bottomrule
\end{tabular}
\caption{Trajectory termination outcomes by benchmark and feedback condition. Percentages are over trajectories within each (benchmark, condition) cell. Under oracle score feedback, correct trajectories terminate immediately upon a fully correct submission, so the balance between repeated similar answers and budget exhaustion is most comparable across benchmarks within, rather than across, feedback conditions. * \textit{Under oracle feedback, HealthBench counts a trajectory "Correct" only when a perfect continuous score of 1.0 is achieved, which never occurred in this evaluation.}
\label{tab:serial-exhaustion}}
\end{table*}

\paragraph{Self-guided iteration often matters less for newer models.}
We also asked whether the value of repeated submission changes across model generations. For each (benchmark, model, condition) cell, we measure iteration uplift as the increase in cumulative success from the first submission to the highest level reached thereafter, and relate it to model release date using Kendall rank correlations ($\tau$; \Cref{fig:serial-cumulative}B; Appendix \Cref{tab:serial-uplift-correlations}). These correlations are descriptive only, as each (benchmark, condition) cell contains just six models. With that caveat, the broad pattern is that newer models tend to find repeated submission less helpful on HLE and TerminalBench, consistent with newer models solving more tasks on their first submission and leaving less room for improvement. On HealthBench, newer models appear to benefit somewhat more (positive $\tau$ in both conditions). SWE-Bench Pro shows essentially no generational effect ($\tau=+0.07$ in both conditions), and arguably neither does FrontierMath, where the two feedback conditions point in opposite directions and estimates rest on only twelve tasks.

\begin{figure*}[!h]
 \centering
 \includegraphics[width=\linewidth]{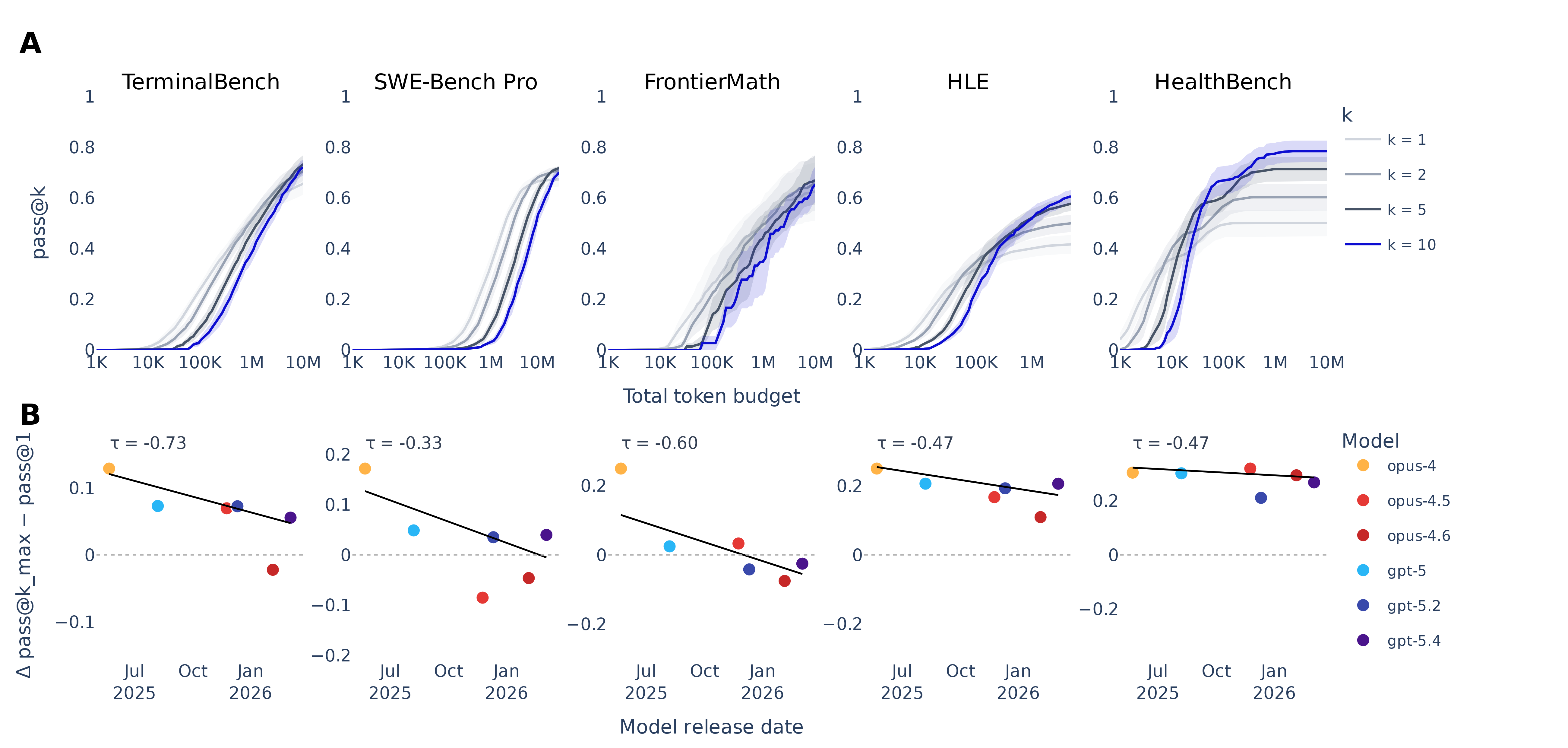}
 \caption{%
  \textbf{Parallel scaling: width vs depth, and uplift across generations.}
  (\textbf{A}) Per benchmark, $x$-axis: total token budget $B$ on a log scale; $y$-axis: $\text{pass@}k$ at total budget $B$ when the budget is split into $k$ trajectories of $B/k$ tokens each. Lines show $k = 1$ (light grey), 2, 5, and 10 (navy). Bands: $\pm 1$ SEM across models.
  (\textbf{B}) Per benchmark, $x$-axis: model release date; $y$-axis: $\Delta_\text{parallel} = \text{pass@}10(B^\star) - \text{pass@}1(B^\star)$. One marker per model. The line is a Theil--Sen trend fit (\Cref{app:parallel-scaling-methods}); the inset reports Kendall $\tau$.
 }
 \label{fig:parallel-scaling}
\end{figure*}

\subsubsection{Serial versus parallel allocation of fixed total compute}
\label{sec:parallel-scaling}

The interventions so far all allocate more compute to a single trajectory -- \emph{serial} scaling. The same total compute can instead be split across multiple independent trajectories. This matters for evaluation because fixed-budget single-trajectory protocols may understate capability either by starving individual runs of depth or by failing to capture gains from independent restarts. We compare pass@$k$ at fixed total budget $B$, splitting $B$ across $k \in \{1,2,5,10\}$ independent trajectories; the exact estimator, pooling details, and our choice of pass@$k$ over consensus-based aggregation are reported in \Cref{app:parallel-scaling-methods}.

\begin{table}[!h]
\centering
\begin{tabular}{lrrrr}
\toprule
Benchmark & pass@1 & pass@10 & $\Delta_\text{parallel}$ & $\tau_\text{gen}$ \\
\midrule
FrontierMath  & 0.625 & 0.653 & $+0.028$ & $-0.60$ \\
HLE      & 0.417 & 0.606 & $+0.189$ & $-0.47$ \\
HealthBench  & 0.501 & 0.784 & $+0.283$ & $-0.47$ \\
SWE-Bench Pro & 0.673 & 0.701 & $+0.028$ & $-0.33$ \\
TerminalBench & 0.657 & 0.720 & $+0.063$ & $-0.73$ \\
\bottomrule
\end{tabular}
\caption{Parallel-sampling gains at each benchmark's token cap. pass@1 and pass@10 are averaged across models. $\Delta_\text{parallel}$ is pass@10 minus pass@1 at the benchmark's token cap. $\tau_\text{gen}$ is the Kendall rank correlation between per-model $\Delta_\text{parallel}$ and release date.}
\label{tab:parallel-gains}
\end{table}

\paragraph{Benchmarks differ in how much they benefit from parallel scaling.}
At each benchmark's token cap, distributing total token budget from a single, deep trajectory to ten shallower trajectories improves average performance on all five benchmarks, but by different amounts: FrontierMath $0.625 \rightarrow 0.653$ ($\Delta_\text{parallel}=+0.028$), HLE $0.417 \rightarrow 0.606$ ($+0.189$), HealthBench $0.501 \rightarrow 0.784$ ($+0.283$), SWE-Bench Pro $0.673 \rightarrow 0.701$ ($+0.028$), and TerminalBench $0.657 \rightarrow 0.720$ ($+0.063$) (\Cref{tab:parallel-gains}; \Cref{fig:parallel-scaling}A). Parallel sampling therefore matters substantially for HLE and HealthBench, but much less for FrontierMath, TerminalBench, and SWE-Bench Pro. We separately verified that the larger gains on the LLM-graded benchmarks (HLE and HealthBench) are not driven by judge unreliability (\Cref{app:scorer-reliability}).

\paragraph{Benefit of parallel sampling depends on budget.}
To assess when width becomes preferable to depth, we assess the total budget $B$ at which pass@$10$ exceeds pass@$1$ (\Cref{fig:parallel-scaling}A). This crossover occurs very early on HealthBench (from roughly 23k total tokens) and HLE (from roughly 219k), but only near the cap on FrontierMath (9M--10M), TerminalBench (5M--10M), and SWE-Bench Pro (23M--30M). These patterns indicate that parallel sampling helps only once each branch receives enough compute to make progress within each trajectory, and that benchmarks differ substantially in how much depth each independent attempt requires before extra width becomes useful.

\paragraph{Newer models generally benefit less from parallel sampling.}
On all five benchmarks, the gain from parallel sampling decreases with model release date (\Cref{tab:parallel-gains}; \Cref{fig:parallel-scaling}B): TerminalBench ($\tau=-0.73$), FrontierMath ($-0.60$), HealthBench ($-0.47$), HLE ($-0.47$), and most weakly SWE-Bench Pro ($\tau=-0.33$). For the newest models, parallel sampling sometimes does not help at all: on FrontierMath, GPT-5.2, Opus 4.6, and GPT-5.4 all achieve lower pass@10 than pass@1 at the cap, and Opus 4.6 shows the same pattern on TerminalBench. Newer models therefore often extract more value from a single long trajectory, reducing the marginal benefit of independent restarts.

\section{Discussion}

\subsection{Summary}

Across seven challenging benchmarks, frontier-model performance improves with additional inference-time compute -- however, the magnitude and shape of those gains differ substantially across settings. Under our protocol, typical published budgets leave meaningful additional performance unrealised on FrontierMath, HLE, and the cyber evaluations, but yield much smaller gains on HealthBench and the two software engineering benchmarks. Cross-generational improvements at larger budgets come mainly from greater reach and reliability rather than from consistently better token efficiency. Measured performance is also sensitive to protocol choices: repeated submission improves performance on all five benchmarks in the main suite, oracle feedback helps most where it enables continued search, and the value of parallel sampling depends strongly on task structure and budget regime. Taken together, these results show that benchmark scores reflect not only the model being evaluated, but also the inference-time budget and protocol used to elicit its performance.

\subsection{Interpretation}

A central question is why responsiveness to the inference-scaling interventions tested here differs so much across benchmarks. These differences should be interpreted as properties of the aggregate performance curves observed under our protocol, not as evidence that any benchmark is intrinsically compute-saturated. Apparent flattening may still conceal headroom under larger budgets, alternative scaffolds, or different allocations of inference-time compute, and benchmark-level curves can mask substantial heterogeneity across tasks.

One plausible interpretation is that the key distinction is not domain alone, but whether a benchmark gives models opportunities to use additional compute productively through extended search, revision, tool use, or self-verification \citep{brown2024monkeys,snell2024scaling,balachandran2025inferencetime}. This view is consistent with the stronger scaling we observe on FrontierMath and the cyber tasks, the more modest but continuing gains on TerminalBench, the smaller gains on SWE-Bench Pro, and the weak scaling on HealthBench under our scaffold. HLE appears intermediate: although stateless, it still benefits considerably from repeated submissions, especially when correctness feedback is available. More broadly, this perspective may help explain why recent work has found strong inference scaling in some settings \citep{folkerts2026cyber,muennighoff2025s1,wu2024inference,ma2025thinking,epoch2026mirrorcode} and weak or absent scaling in others \citep{gema2025inverse,jurkovic2026metr,terminalbench2026}.

Our cross-generational results suggest that recent model progress has improved the ability to use inference-time compute productively on tasks that permit it. Later models usually achieve higher performance at large budgets and often begin succeeding earlier in the token allowance, but these gains come mainly from increased reach and reliability rather than from large improvements in token efficiency. This pattern is consistent with recent time-horizon analyses attributing frontier-model progress primarily to greater reliability and the ability to tackle longer or harder tasks \citep{kwa2025timehorizon,metr2026timehorizon11}. It also implies that fixed-budget evaluations may become increasingly uneven in what they capture, understating progress most on tasks that benefit from extended search, verification, or repeated attempts.

The submission-protocol and parallel-sampling results reinforce the same broader conclusion. Repeated submissions improved performance on all five main benchmarks, oracle feedback helped most where it supported continued search, and parallel sampling helped most when tasks admitted multiple viable independent attempts. This is broadly consistent with prior work on width--depth trade-offs, external verification, and repeated search in LLM evaluation and elicitation \citep{brown2024monkeys,snell2024scaling,huang2024selfcorrect,stechly2024selfverification,balachandran2025inferencetime}. More generally, inference-time compute, feedback conditions, submission rules, and width--depth allocation should be treated as part of the measurement definition rather than as incidental implementation details.

\subsection{Recommendations}

\paragraph{Report capability over a range of inference-time compute, not just at a single budget} 
A single fixed-budget score can either approximate high-budget performance reasonably well or omit substantial additional gains, depending on the benchmark. Reporting performance as a function of inference-time compute helps distinguish near-saturated evaluations from budget-censored lower bounds, and clarifies when additional compute continues to buy measurable performance. This is especially important when evaluations are used to characterise what well-resourced actors could achieve, rather than only what is accessible under restrictive default settings \citep{cerruti2026safety_eval_ttc}. Where possible, evaluators should also distinguish whether gains come from solving new tasks, solving the same tasks more reliably, or solving already-reachable tasks with less compute, since aggregate scaling curves can obscure these differences.

\paragraph{Treat protocol choices as part of the evaluation design and clearly document them}
In this study, repeated submissions, feedback after submission, and the allocation of compute across serial depth or parallel width all meaningfully changed measured performance on at least some benchmarks. These choices should therefore be reported explicitly and justified relative to the goals of the evaluation, the deployment setting, or the actor scenario the evaluation is meant to capture. Oracle correctness feedback is most ecologically valid where external verifiers already exist, such as code execution, formal proofs, or grounded tool outputs, and less so for open-ended expert reasoning tasks. Likewise, some benchmarks benefited much more from parallel sampling than others, implying that the same total compute can elicit meaningfully different levels of performance depending on how it is allocated.

\paragraph{For trend tracking, compare models at matched budgets within a large, shared compute range} 
Newer models often reach higher performance only at larger budgets, and they do not always use additional inference-time compute in the same way as earlier generations. In our task-level analyses, generational gains came mainly from greater reach and reliability rather than from consistently improved token efficiency. Evaluations used for trend tracking, long-horizon performance estimation, or safety thresholding should therefore compare models under the same protocol and over the same broad compute range, or explicitly account for differences in those choices. Otherwise, fixed-budget comparisons may become increasingly misleading as models improve, especially on tasks where performance depends strongly on extended search, verification, or repeated attempts.

\subsection{Limitations}

First, we study a single ReAct-style scaffold, even though bespoke scaffolds and elicitation effort can substantially shift measured capability on the same benchmark and model \citep{metr2024elicitation,aisi2024earlylessons,jurkovic2026metr,sun2025contextfolding}. Some benchmark differences we observe may therefore be scaffold-dependent rather than task-intrinsic, and the weak scaling on HealthBench in particular may reflect genuine knowledge-boundedness, judge saturation, scaffold mismatch, or some combination \citep{ravichandran2025drinfo}. 

Second, our protocol is deliberately simple -- a general combination of larger token budgets, context compaction, and iterative resubmission, but not more elaborate strategies such as adaptive branching, verifier-guided search, or hybrid serial--parallel allocation within and across trajectories. Such methods may materially change absolute performance, possibly differently across domains, so our results should be interpreted as a conservative lower bound on what broadly reproducible inference scaling can achieve rather than an upper bound under more optimised elicitation. 

Third, FrontierMath contains only twelve tasks in our subset, reducing power for the task-level analyses and making benchmark-level patterns sensitive to individual items; newer models may additionally have training overlap with the public set \citep{burnham2025frontiermath}. 

Fourth, the repetition guard relies on an LLM judge, whose reliability at detecting semantic equivalence may vary across benchmarks, making the guard's effective strictness non-uniform across domains.

\bibliographystyle{unsrtnat}
\bibliography{references}

\newpage
\appendix
\section{Appendix}

\subsection{Experimental protocol}

\subsubsection{Benchmark details}
\label{app:benchmark-details}

This section gives per-benchmark details on dataset loading, filtering, and grading. All benchmarks are loaded via Inspect AI-compatible wrappers, and all LLM-graded scoring uses GPT-4o-mini at temperature 0. For stateful benchmarks, tool calls that exceed 180 seconds are terminated and return an error to the model, which may then continue to the next turn.

\paragraph{TerminalBench.}
We use the Terminal-Bench 2.0 release \citep{terminalbench2026} accessed via the \texttt{inspect\_harbor} registry, restricted to the 86 tasks, out of 89, whose scorer reliably emits a score event. Each task ships with its own pre-built Docker image, with CPU, memory, and GPU specifications defined per task, and verification is performed by the task's bundled test harness with binary pass/fail output. Tasks are executed in a Kubernetes-backed sandbox, one pod per trajectory, using \texttt{inspect-k8s-sandbox}, with per-task images pulled from a private ECR registry.

\paragraph{SWE-Bench Pro.}
We use SWE-Bench Pro \citep{swebenchpro2025} via the \texttt{inspect\_harbor} registry. The registry ships two prompt variants per task, 1{,}462 samples across 731 unique instances. For comparability with published baselines, we keep a single variant per task, the SWE-bench-format variant whose input begins with \texttt{<uploaded\_files>}, yielding 731 unique problems. A canonical subset of the first 100 problems is drawn deterministically and fixed across all runs. Verification uses each instance's bundled test harness against the candidate patch with binary pass/fail output. Per-task Docker images are pre-built and pushed to a private ECR registry, and trajectories run in Kubernetes-backed sandboxes.

\paragraph{FrontierMath.}
We use the public dataset for FrontierMath \citep{glazer2024frontiermath}, loaded from a local JSONL file. Each record contains the question text, the expected return type, and a per-problem verification code block. Solutions are submitted as Python functions, an \texttt{answer()} function that takes no arguments and returns a value of the declared type, with no comments or \texttt{print} output. The scorer runs the submitted \texttt{answer()} followed by the problem's \texttt{verify(a)} in a sandboxed Python environment and returns CORRECT or INCORRECT based on the verifier's exit code. The \texttt{answer()} function has a 30-second execution timeout, and the verification code has a 120-second timeout. The available Python libraries are \texttt{sympy}, \texttt{numpy}, \texttt{scipy}, \texttt{mpmath}, \texttt{gmpy2}, \texttt{pyadic}, \texttt{galois}, and \texttt{networkx}. The full pinned version set is distributed with the sample via the system prompt. The sandbox is a self-managed Docker compose stack, one container per trajectory.

\paragraph{HealthBench.}
We use HealthBench \citep{openai2025healthbench} via its \texttt{inspect\_evals} wrapper, restricted to the subset of hard samples, ``HealthBench Hard''. Each sample's reference is a physician-designed rubric, a list of criteria, each with an associated integer point value, positive for desirable behaviours and negative for penalised ones. Scoring proceeds by evaluating each criterion independently with an LLM judge, GPT-4o-mini at temperature 0, which returns a boolean judgment. The sample's continuous score is the sum of the points of met criteria divided by the sum of positive points. This generally lies in $[0, 1]$ but can fall below zero when negative-weighted criteria are triggered. No tools or sandbox are required.

\paragraph{HLE.}
We use Humanity's Last Exam \citep{phan2025hle} via its \texttt{inspect\_evals} wrapper. We filter the dataset to exact-match questions only, identified by the presence of the phrase ``Answer:'' or ``Exact Answer:'' in the sample's system message, excluding multiple-choice questions that would otherwise allow an iterative agent to guess its way to the correct answer. Grading is performed by an LLM judge, GPT-4o-mini at temperature 0, using HLE's provided judge prompt. The judge is directed to focus on whether the response is semantically equivalent to the reference answer and to emit a final verdict in the format \texttt{GRADE: C} (correct) or \texttt{GRADE: I} (incorrect). Where available, a ``Confidence'' tag reported by the model is also recorded as metadata. No tools or sandbox are required.

In addition to the five main benchmarks that we ran directly for this study, we include previously collected results on two cybersecurity benchmarks, Cyber CTFs and The Last Ones \citep{aisi2026mythos,folkerts2026cyber,aisi2026scaling}. These data were not re-run under the fully crossed design of the present paper. We include them only in the inference-scaling analyses because their collection protocol is closely comparable on the dimensions relevant here, namely large inference budgets, repeated agent interaction in externally verifiable environments, context compaction, and multiple independent trajectories per task.

\paragraph{Cyber CTFs.}
Cyber CTFs is a suite of 71 isolated capture-the-flag tasks spanning common offensive-security skill areas such as web exploitation, cryptography, and reverse engineering \citep{aisi2026mythos,aisi2026scaling}. Each task is scored by binary flag capture. In the reused data, models were evaluated with five independent trajectories per task under a high-budget, externally verifiable protocol with a 50M total-token cap, and only the final submission per trajectory is retained for scoring.

\paragraph{The Last Ones.}
The Last Ones is a single long-horizon cyber range task representing a simulated corporate network attack \citep{folkerts2026cyber}. The objective is to exfiltrate sensitive data from a protected internal database by chaining together reconnaissance, exploitation, credential theft, lateral movement, reverse engineering, and subsequent access-establishment steps across a multi-host environment. The attack chain contains 32 steps grouped into 9 milestones, and progress is scored continuously by milestones toward the final objective. In the reused data, each in-trajectory milestone reached is treated as a submission event for inference-scaling purposes. Runs use five independent trajectories and large total token budgets, up to 100M, with context compaction used to support long trajectories. The original evaluation employed a standard ReAct-style agent with access to Kali Linux tooling and additional cyber-specific tooling in a virtualised environment \citep{folkerts2026cyber}.

\subsubsection{Typical published budget estimates}
\label{app:typical-budgets}

To compare our expanded inference-time budgets against standard evaluation practice, we define a benchmark-specific ``typical'' budget for each benchmark. Because published evaluations use different stopping rules, we convert all reported limits into approximate per-trajectory total-token budgets.

For three benchmarks, we retrieved specific per-trajectory token limits from public sources. These are 16--64k output tokens for HLE \citep{phan2025hle}, 16k output tokens for HealthBench \citep{openai2025healthbench}, and 1M total tokens for FrontierMath as reported by Epoch AI \citep{burnham2025frontiermath}. The latter is a tenfold increase over their earlier 100k-token scaffold, while the original preprint does not specify a fixed per-trajectory budget \citep{glazer2024frontiermath}.

TerminalBench and SWE-Bench Pro leaderboards do not specify token limits. TerminalBench uses per-task wall-clock limits of 10 minutes to 3.3 hours, with a median of 15 minutes \citep{terminalbench2026}. SWE-Bench Pro uses turn caps of 50--250 \citep{swebenchpro2025,scale2026sealswebenchpro}. To convert these to token equivalents, we use our own collected trajectories. For each trajectory, we record the cumulative total tokens consumed by the point at which the leaderboard's wall-clock or turn limit would have applied. We average these totals within each model over tasks and trajectories and take the median across models. This yields 1M--7.3M tokens for TerminalBench and 473k--16M tokens for SWE-Bench Pro. These estimated token limits are displayed in \Cref{fig:scaling-curves}A to illustrate inference-scaled uplift. A full per-benchmark, per-source tabulation of reported scores and stopping limits is given in Appendix \Cref{tab:eval-setups}. The final typical budget estimates per benchmark are reported in \Cref{tab:budget-uplift}.

\begin{landscape}
\footnotesize
\setlength{\tabcolsep}{4pt}
\begin{longtable}{>{\raggedright\arraybackslash}p{2.3cm}>{\raggedright\arraybackslash}p{2cm}>{\raggedright\arraybackslash}p{3.5cm}>{\raggedright\arraybackslash}p{1.6cm}>{\raggedright\arraybackslash}p{1.0cm}>{\raggedright\arraybackslash}p{1.7cm}>{\raggedright\arraybackslash}p{1.4cm}>{\raggedright\arraybackslash}p{1.9cm}>{\raggedright\arraybackslash}p{1.6cm}>{\raggedright\arraybackslash}p{1.2cm}>{\raggedright\arraybackslash}p{1.8cm}}
\caption{Reported evaluation setups in prior work for each benchmark studied in this paper. One row per (benchmark, source citation), aggregating across 89 distinct harness setups documented in 61 citations; where a single citation documents multiple setups, each parameter value is shown as the range across those setups (e.g. ``10M--100M''). Sources are restricted to benchmark preprints, lab system / model cards, and prominent third-party leaderboards. Across the 60 citations, stopping limits are undocumented at the following rates: token budget 42/61 (69\%), turn cap 57/61 (93\%), wall-clock cap 47/61 (77\%), submissions per trajectory 28/61 (46\%), cost cap 57/61 (93\%), trajectories per task 34/61 (56\%). An em-dash (---) indicates the citation did not disclose that field in any setup.} \label{tab:eval-setups} \\
\toprule
Benchmark & Citation & Source title & Source type & Date & Token limit & Turn cap & Time limit & Submissions / traj. & Cost cap & Independent traj. / task \\
\midrule
\endfirsthead
\multicolumn{11}{l}{\footnotesize\emph{Table~\ref{tab:eval-setups} continued.}} \\
\toprule
Benchmark & Citation & Source title & Source type & Date & Token limit & Turn cap & Time limit & Submissions / traj. & Cost cap & Independent traj. / task \\
\midrule
\endhead
\midrule
\multicolumn{11}{r}{\footnotesize\emph{continued on next page}} \\
\endfoot
\bottomrule
\endlastfoot
TerminalBench 2.0 & \citeevals{google2026gemini31pro} & Google DeepMind Gemini 3.1 Pro model card & system card & 2026-05? & --- & --- & --- & --- & --- & --- \\
TerminalBench 2.0 & \citeevals{tbenchleaderboard} & tbench.ai leaderboard submission rules & leaderboard & 2026-05? & --- & --- & per-task (benchmark default; not modifiable) & 1 & --- & 5 \\
TerminalBench 2.0 & \citeevals{anthropic2026mythos} & Claude Mythos Preview release (Glasswing) & blog post & 2026-04? & 1M per task & --- & per-task (benchmark default; up to 4h on TB 2.1) & 1 & --- & 5 \\
TerminalBench 2.0 & \citeevals{anthropic2026opus47} & Claude Opus 4.7 announcement & blog post & 2026-04? & --- & --- & per-task (benchmark default) & 1 & --- & 5 \\
TerminalBench 2.0 & \citeevals{anthropic2026sonnet46card} & Claude Sonnet 4.6 system card, section 2.3 & system card & 2026-04? & --- & --- & per-task (benchmark default) & 1 & --- & 5 \\
TerminalBench 2.0 & \citeevals{anthropic2026opus46} & Claude Opus 4.6 announcement & blog post & 2026-03? & --- & --- & per-task (benchmark default) & 1 & --- & 5-15 \\
TerminalBench 2.0 & \citeevals{openai2026gpt54} & OpenAI GPT-5.4 release & blog post & 2026-03? & --- & --- & --- & --- & --- & --- \\
TerminalBench 2.0 & \citeevals{openai2026gpt53codex} & OpenAI GPT-5.3 Codex release ("Simple Codex" harness) & blog post & 2026-02? & --- & --- & --- & --- & --- & --- \\
TerminalBench 2.0 & \citeevals{terminalbench2026} & Terminal-Bench 2.0 paper, Appendices A \& F (Terminus-2 harness) & preprint & 2026-01 & no cap & no hard cap & per-task (default \textasciitilde 6 min, max 2h) & 1 & --- & 5 \\
\midrule
SWE-Bench Pro & \citeevals{llmstats2026swebenchpro} & LLM-Stats SWE-Bench Pro leaderboard & leaderboard & 2026-05 & --- & --- & --- & --- & --- & --- \\
SWE-Bench Pro (public) & \citeevals{scale2026sealswebenchpro} & Scale SEAL leaderboard & leaderboard & 2026-05 & --- & 50--250 & --- & 1 & --- & 1 \\
SWE-Bench Pro (public) & \citeevals{alibaba2026qwen36max} & Alibaba Qwen 3.6 Max Preview launch & blog post & 2026-04 & --- & --- & --- & --- & --- & --- \\
SWE-Bench Pro (public) & \citeevals{alibaba2026qwen36plus} & Alibaba Qwen 3.6 Plus report & blog post & 2026-04 & --- & --- & --- & --- & --- & --- \\
SWE-Bench Pro (public) & \citeevals{anthropic2026mythos} & Anthropic Claude Mythos Preview (Project Glasswing) & blog post & 2026-04 & --- & --- & --- & --- & --- & --- \\
SWE-Bench Pro (public) & \citeevals{anthropic2026opus47} & Anthropic Claude Opus 4.7 release & blog post & 2026-04 & --- & --- & --- & --- & --- & --- \\
SWE-Bench Pro (public) & \citeevals{augment2026auggie} & Augment Auggie blog (Opus 4.5 on Auggie) & blog post & 2026-04 & --- & --- & --- & --- & --- & --- \\
SWE-Bench Pro (public) & \citeevals{blitzy2026swebenchpro} & Blitzy SWE-Bench Pro audit & blog post & 2026-04 & --- & --- & --- & 1 & --- & --- \\
SWE-Bench Pro (public) & \citeevals{cursor2026composer} & Cursor Composer 2 release & blog post & 2026-04 & --- & --- & --- & --- & --- & --- \\
SWE-Bench Pro (public) & \citeevals{minimax2026m27} & MiniMax M2.7 launch & blog post & 2026-04 & --- & --- & --- & --- & --- & --- \\
SWE-Bench Pro (public) & \citeevals{moonshot2026kimik26} & Moonshot Kimi K2.6 tech blog & blog post & 2026-04 & --- & --- & --- & --- & --- & --- \\
SWE-Bench Pro (public) & \citeevals{morph2026warpgrep} & Morph WarpGrep v2 self-reported & blog post & 2026-04 & --- & --- & --- & --- & --- & --- \\
SWE-Bench Pro (public) & \citeevals{zhipu2026glm51} & Zhipu GLM-5.1 release (Z.ai) & blog post & 2026-04 & 32k output & --- & --- & --- & --- & --- \\
SWE-Bench Pro (public) & \citeevals{google2026gemini31pro} & Google DeepMind Gemini 3.1 Pro model card & system card & 2026-03 & --- & --- & --- & 1 & --- & 1 \\
SWE-Bench Pro (public) & \citeevals{openai2026gpt54} & OpenAI GPT-5.4 release & blog post & 2026-03 & --- & --- & --- & --- & --- & --- \\
SWE-Bench Pro (public) & \citeevals{anthropic2026opus46} & Anthropic Claude Opus 4.6 release & blog post & 2026-02 & --- & --- & --- & --- & --- & --- \\
SWE-Bench Pro (public) & \citeevals{openai2026gpt53codex} & OpenAI GPT-5.3-Codex release & blog post & 2026-02 & --- & --- & --- & --- & --- & --- \\
SWE-Bench Pro & \citeevals{openai2026gpt52} & OpenAI GPT-5.2 release & blog post & 2026-01 & --- & --- & --- & --- & --- & --- \\
SWE-Bench Pro (public) & \citeevals{windsurf2025swe15} & Windsurf/Cognition SWE-1.5 launch & blog post & 2025-10 & --- & --- & --- & --- & --- & --- \\
SWE-Bench Pro (public, commercial) & \citeevals{swebenchpro2025} & SWE-Bench Pro paper & preprint & 2025-09 & --- & 50 & --- & 1 & \$2 & 1 \\
\midrule
FrontierMath (tier 4) & \citeevals{epoch2026gpt52pro} & Epoch AI substack (GPT-5.2 Pro Tier 4 record) & blog post & 2026-01 & --- & --- & --- & --- & --- & 1 \\
FrontierMath (tiers 1-3) & \citeevals{openai2025gpt52blog} & OpenAI GPT-5.2 announcement (science \& math) & blog post & 2025-12 & --- & --- & --- & 1 & --- & 1 \\
FrontierMath (full) & \citeevals{anthropic2025opus45} & Anthropic Claude Opus 4.5 announcement & blog post & 2025-11 & --- & --- & --- & --- & --- & --- \\
FrontierMath (tiers 1-3) & \citeevals{burnham2025frontiermath} & Epoch AI 1M-token scaffold sweep (Burnham) & blog post & 2025-11 & 1M & --- & 30s/tool call & 1 & --- & 32 \\
FrontierMath (tiers 1-3) & \citeevals{epoch2025gemini3} & Epoch AI X post (Gemini 3 Pro) & blog post & 2025-11 & 1M & --- & 30s/tool call & 1 & --- & 1 \\
FrontierMath (tiers 1-3) & \citeevals{epoch2025gpt51} & Epoch AI benchmarks hub (GPT-5.1 era) & leaderboard & 2025-11 & 1M & --- & 30s/tool call & 1 & --- & 1 \\
FrontierMath (tiers 1-3) & \citeevals{epoch2025opus45} & Epoch AI X post (Opus 4.5) & blog post & 2025-11 & 1M & --- & 30s/tool call & 1 & --- & 1 \\
FrontierMath (tiers 1-3) & \citeevals{epoch2025grok4math} & Epoch AI Grok 4 math eval blog & blog post & 2025-07 & 100k & --- & 30s/tool call & 1 & --- & 1 \\
FrontierMath (tier 4) & \citeevals{epoch2025tier4} & Epoch AI Tier 4 benchmark page & leaderboard & 2025-07 & 1M & --- & 30s/tool call & 1 & --- & 1 \\
FrontierMath (tiers 1-3) & \citeevals{epoch2025o4mini} & Epoch AI scaffold v1.0 (o4-mini eval) & blog post & 2025-04 & 100k & --- & 30s/tool call & 1 & --- & 1 \\
FrontierMath (tiers 1-3) & \citeevals{epoch2025o3mini} & Epoch AI scaffold v1.0 (o3-mini era) & blog post & 2025-03 & 100k & --- & 30s/tool call & 1 & --- & 1 \\
FrontierMath (full) & \citeevals{glazer2024frontiermath} & FrontierMath paper (Glazer et al.) & preprint & 2024-11 & 10k conversation & --- & --- & 1 & --- & 1 \\
\midrule
HealthBench (full) & \citeevals{openai2026gpt54card} & GPT-5.4 Thinking system card & system card & 2026-03 & --- & --- & --- & 1 & --- & --- \\
HealthBench (full, Hard) & \citeevals{openai2025gpt52card} & GPT-5.2 system card, Section 3.6 & system card & 2025-12 & --- & --- & --- & 1 & --- & --- \\
HealthBench (full, Hard, Consensus) & \citeevals{openai2025gpt5card} & GPT-5 system card & system card & 2025-08 & --- & --- & --- & 1 & --- & --- \\
HealthBench (full, Hard, Consensus) & \citeevals{openai2025healthbench} & HealthBench paper & preprint & 2025-05 & 4096 output, 16k output, 9k output & --- & --- & 1 & --- & --- \\
\midrule
HLE (text-only) & \citeevals{scale2025hletextonly} & Scale SEAL HLE text-only leaderboard methodology & leaderboard & 2026-05 & --- & --- & --- & 1 & --- & --- \\
HLE (full) & \citeevals{scale2026hleleaderboard} & Scale SEAL HLE leaderboard methodology & leaderboard & 2026-05 & --- & --- & --- & 1 & --- & --- \\
HLE (full) & \citeevals{google2025gemini3} & Google Gemini 3 launch & blog post & 2025-11 & --- & --- & --- & 1 & --- & --- \\
HLE (full) & \citeevals{google2025gemini3deepthink} & Google Gemini 3 Deep Think blog & blog post & 2025-11 & --- & --- & --- & 1 & --- & --- \\
HLE (full) & \citeevals{moonshot2025kimik2thinking} & Kimi K2 Thinking & system card & 2025-11 & 96k thinking, 48k per step & 120 & --- & 1 & --- & 8 \\
HLE (full) & \citeevals{openai2025gpt5} & OpenAI GPT-5 system card & system card & 2025-08 & --- & --- & --- & 1 & --- & --- \\
HLE (text-only) & \citeevals{xai2025grok4launch} & xAI Grok 4 & blog post & 2025-07 & --- & --- & --- & 1 & --- & 4 \\
HLE (full) & \citeevals{openai2025deepresearch} & OpenAI Deep Research launch & blog post & 2025-02 & --- & --- & --- & 1 & --- & --- \\
HLE (full, text-only) & \citeevals{phan2025hle} & HLE & preprint & 2025-01 & 8192 output & --- & --- & 1 & --- & --- \\
\midrule
Cyber CTFs (apprentice tier, practitioner/expert tier) & \citeevals{aisi2026scaling} & AISI inference scaling blog & blog post & 2026-04? & 2.5M--50M & --- & --- & --- & \$60 & 10 \\
Cyber CTFs (expert-level, apprentice/non-expert) & \citeevals{aisi2026mythos} & AISI Mythos Preview cyber capabilities & blog post & 2026-03? & 2.5M--50M & --- & --- & --- & --- & 5--10 \\
Cyber CTFs (71 challenges) & \citeevals{folkerts2026cyber} & Folkerts et al. 2026, Appendix D (AISI CTF suite) & preprint & 2026-03 & --- & --- & --- & --- & --- & 10 \\
Cyber CTFs (47 challenges) & \citeevals{aisi2024o1} & AISI pre-deployment eval of OpenAI o1 & blog post & 2024-12 & --- & --- & --- & --- & --- & --- \\
Cyber CTFs (47 challenges) & \citeevals{aisi2024sonnet35} & AISI pre-deployment eval of Claude 3.5 Sonnet (upgraded) & blog post & 2024-10 & --- & --- & --- & --- & --- & --- \\
\midrule
AISI The Last Ones & \citeevals{folkerts2026cyber} & Folkerts et al. 2026, Table 1 & preprint & 2026-03 & 10M--100M & --- & --- & --- & \textasciitilde \$80 & 5--10 \\
\end{longtable}

\vspace{0.5em}
\noindent\rule{0.3\linewidth}{0.4pt}
\begingroup
\footnotesize
\setlength{\bibsep}{0pt plus 0.3ex}
\renewcommand{\bibsection}{}
\bibliographyevals{references}
\endgroup
\end{landscape}

\subsubsection{Repetition guard}
\label{app:repetition-guard}

The repetition guard terminates a trajectory early when the agent appears stuck in an unproductive submission loop. Whenever the agent issues a submission, the guard identifies the maximal suffix of consecutive agent turns that contain \emph{only} submit tool calls, with no intervening \texttt{bash} or \texttt{python} invocations. If this suffix contains three or more submissions, the guard sends the full set of submissions to an LLM judge, GPT-4o-mini at temperature 0, and asks whether they are semantically equivalent. On a \emph{SAME} verdict, the trajectory is marked as stopped by the repetition guard and terminated. On \emph{DIFFERENT}, execution continues normally. For FrontierMath specifically, interleaved non-submit tool calls are permitted during the backward scan. In other words, the guard considers all submissions in the trajectory rather than only the current submit-only suffix. This avoids penalising long sequences of exploratory computation between submissions. The guard is applied live during evaluation for most runs and retroactively, with a cached judge, during analysis for any runs that were missed.

\subsubsection{Experimental conditions: prompts and feedback}
\label{app:conditions-prompts}

Both conditions share a single adaptive \emph{continuation} prompt, appended after every non-terminal submission to invite the agent either to refine its previous answer or to try a substantially different approach:

\begin{quote}
\emph{``Consider whether your previous approach is likely correct or whether a different approach would be more promising. Then either refine your previous answer or try a substantially different approach, and submit when ready.''}
\end{quote}

The per-submission feedback strings are defined in the main text. The following protocol details supplement those definitions:

\begin{itemize}
\item \textbf{No feedback.} The acknowledgement is identical on correct and incorrect submissions, giving the agent no implicit signal of correctness.
\item \textbf{Oracle score feedback (binary benchmarks: TerminalBench, SWE-Bench Pro, FrontierMath, HLE).} On a correct submission no feedback message is sent, and the trajectory simply ends.
\item \textbf{Oracle score feedback (continuous benchmark: HealthBench).} The full feedback string is \emph{``Your submission was scored. You achieved $X$ out of $Y$ possible points.''}, where $X$ is the net points achieved on the most recent submission and $Y$ is the sum of positive-weighted points for the sample. Trajectories terminate only when the agent achieves every positively weighted criterion. Partial-credit submissions trigger the feedback message followed by the adaptive continuation.
\end{itemize}

\subsection{Inference-scaling curves: supplementary analyses}

\subsubsection{Curve-level cross-generational summaries}
\label{app:curve-characteristics}

To summarise cross-generational patterns at the curve level, we extract three descriptive quantities from each pooled benchmark--model curve:

\begin{itemize}
    \item \textbf{Success at the maximum tested budget} is the highest performance reached within our observed range, that is, the curve's right-edge value in \Cref{fig:scaling-curves}A.
    \item \textbf{Onset point} is the minimum compute level at which a model first begins to solve tasks, measured as the first total-token budget at which cumulative success exceeds 5\%.
    \item \textbf{Post-onset slope} measures how strongly performance continues to improve after onset. We compute it as the average increase in cumulative success per tenfold increase in total tokens from onset to the point where the curve begins to flatten, as defined in \Cref{sec:plateau-results}. If a curve does not flatten within the tested range, we use the maximum tested budget instead.
\end{itemize}

\paragraph{Success at the maximum tested budget.} Across benchmarks, the clearest and most consistent pattern is that later-release models reach higher performance at the maximum tested budget (\Cref{tab:curve_characteristics}). The Kendall rank correlation between model release date and success at the cap is strongly positive on Cyber CTFs ($\tau=0.944$), TerminalBench ($\tau=0.867$), FrontierMath ($\tau=0.867$), and HLE ($\tau=0.733$), and remains positive though weaker on SWE-Bench Pro ($\tau=0.414$). HealthBench is the main exception, showing little relationship between release date and success at the cap ($\tau=0.067$). On most benchmarks in our study, newer models therefore reach a higher protocol-conditional performance frontier when given large inference-time budgets.

\paragraph{Onset.} Later-release models also often begin succeeding earlier in the budget range. On every benchmark, onset is negatively correlated with release date, with $\tau$ ranging from $-0.333$ to $-0.467$. Newer models often need fewer total tokens before they start to solve tasks at all.

\paragraph{Post-onset slope.} Whether newer models also use \emph{additional} compute better after that point is more benchmark-dependent. Post-onset slope increases clearly with model recency on Cyber CTFs ($\tau=0.944$), FrontierMath ($\tau=0.600$), and TerminalBench ($\tau=0.600$), and remains weak on HLE ($\tau=0.067$), SWE-Bench Pro ($\tau=0.200$), and HealthBench ($\tau=-0.067$). The extent to which newer models convert extra inference-time compute into further gains therefore depends on the task setting.

\begin{table}[!h]
\centering
\begin{tabular}{lccc}
\toprule
Benchmark & Success at maximum tested budget $\tau$ & Onset $\tau$ & Post-onset slope $\tau$ \\
\midrule
Cyber CTFs  & $+0.944$ & $-0.360$ & $+0.944$ \\
TerminalBench & $+0.867$ & $-0.467$ & $+0.600$ \\
SWE-Bench Pro & $+0.414$ & $-0.333$ & $+0.200$ \\
FrontierMath & $+0.867$ & $-0.333$ & $+0.600$ \\
HLE      & $+0.733$ & $-0.467$ & $+0.067$ \\
HealthBench  & $+0.067$ & $-0.467$ & $-0.067$ \\
\bottomrule
\end{tabular}
\caption{Kendall rank correlations between model release date and three descriptive summaries of each pooled inference-scaling curve. Success at cap is cumulative success at the maximum tested budget. Onset point is the first total-token budget at which cumulative success exceeds 5\%. Post-onset slope is the average gain in cumulative success per tenfold increase in tokens between onset and plateau onset, or the tested cap if no plateau is observed.}
\label{tab:curve_characteristics}
\end{table}

\subsection{Task-level decomposition: regression methods and sensitivity analyses}

This appendix gives the regression specifications and a sensitivity check behind the task-level decomposition of generational gains reported in \Cref{sec:unlocks}. The decomposition splits each (model, task) outcome into three components --- reach, efficiency, and reliability --- and estimates each by regressing the relevant outcome on model generation, leave-one-model-out task difficulty, and their interaction. We first define the per-cell outcome measures and predictors, then give each component's regression specification and the coefficient-interpretation conventions, and finally report a balanced-panel sensitivity check for the efficiency analysis (\Cref{app:efficiency-balanced-panel}).

\subsubsection{Outcome measures and predictors}
\label{app:task-level-analysis}

Let $m$ index models and $t$ index tasks.

\paragraph{Per-cell outcome measures.}
For each (benchmark, model, task) cell, we aggregate over all of that model's independent trajectories to construct three quantities.

\begin{itemize}[leftmargin=*, itemsep=2pt, topsep=2pt]
\item \textbf{Solve rate} $s_{m,t}$, the fraction of trajectories containing at least one solving submission. For the five binary benchmarks, a submission is counted as solving if it receives score 1. For HealthBench, which is continuously scored, submissions are first binarised at 0.38, the global median of pooled per-trajectory maximum scores across all HealthBench trajectories.
\item \textbf{Median tokens-to-solve} $\kappa_{m,t}$, the median output token count among solving trajectories only. Within each (benchmark, model) cell, solving trajectories with $|z|>3$ on $\log_{10}(\text{tokens-to-solve})$ are dropped before taking the median to reduce the influence of extreme right-tail outliers.
\item \textbf{Unlock indicator} $u_{m,t}$, an indicator equal to 1 if at least two of the independent trajectories solve the task and 0 otherwise:
\[
u_{m,t} = \mathbb{1}\{n_{\mathrm{solves}} \ge 2\}.
\]
This threshold is intended to distinguish tasks a model can solve reproducibly from those passed only on a single lucky trajectory.
\end{itemize}

\paragraph{Task difficulty.}
We define task difficulty using the cross-model pass rate
\[
\pi_t = \mathrm{median}_m\, s_{m,t},
\]
computed over all models evaluated on task $t$, so that higher values denote easier tasks. For the reliability regression, we use a leave-one-model-out variant
\[
\pi_{-m,t} = \mathrm{median}_{m' \neq m} s_{m',t},
\]
which excludes the focal model's own solve rate from the predictor. These difficulty measures lie in $[0,1]$.

\paragraph{Generation coding.}
For the regression analyses, generation is defined from model release date. Within each benchmark, release dates are min--max normalised to $[0,1]$ and then mean-centred:
\[
g_m = \mathrm{center}\!\left(\mathrm{minmax}(\text{release date}_m)\right).
\]
Difficulty predictors are likewise mean-centred within benchmark before fitting, so that main effects are interpretable at average benchmark difficulty. The reach regression uses these same centred predictors.

\subsubsection{Reach, efficiency, and reliability regressions}

\paragraph{Reach analysis.}
For each benchmark, we fit a linear probability model regressing $u_{m,t}$ on model generation, leave-one-model-out task difficulty, and their interaction, by ordinary least squares (OLS) with cluster-robust standard errors at the task level. In Wilkinson notation,
\[
\texttt{unlocked} \sim \texttt{generation} * \texttt{task\_pass\_rate},
\]
or equivalently
\begin{equation*}
u_{m,t} = \alpha + \beta_{\mathrm{gen}} g_m + \beta_{\mathrm{pass}} \pi_{-m,t} + \beta_{\mathrm{int}}\, g_m\pi_{-m,t} + \varepsilon_{m,t},
\end{equation*}
fit with \texttt{statsmodels.formula.api.ols}. As in the reliability analysis, the leave-one-model-out difficulty predictor $\pi_{-m,t}$ excludes the focal model's own solves, which determine $u_{m,t}$. \Cref{fig:task-level-decomposition}A complements this with the per-benchmark difficulty frontier: the running maximum task difficulty unlocked as a function of the first-unlock model's release date.

\paragraph{Efficiency analysis.}
Efficiency asks whether later generations solve already-unlocked tasks with fewer tokens. This analysis is therefore restricted to unlocked (model, task) cells, where $u_{m,t}=1$ and $\kappa_{m,t}$ is defined. For each benchmark, we regress log median tokens-to-solve on model generation, task difficulty, and their interaction, with a task-level random intercept. In Wilkinson notation,\footnote{In Wilkinson notation, \texttt{a * b} expands to \texttt{a + b + a:b}, that is, both main effects plus their interaction. The trailing \texttt{(1 | task)} denotes a per-task random intercept.}
\[
\texttt{log\_tokens} \sim \texttt{generation} * \texttt{task\_pass\_rate} \; + \; (1 \mid \texttt{task}),
\]
or equivalently
\begin{equation*}
\log_{10}\kappa_{m,t} = \alpha + \beta_{\mathrm{gen}} g_m + \beta_{\mathrm{pass}} \pi_t + \beta_{\mathrm{int}}\, g_m\pi_t + u_t + \varepsilon_{m,t},
\qquad
u_t \sim \mathcal{N}(0,\sigma_u^2),
\end{equation*}
fit with \texttt{statsmodels.formula.api.mixedlm} grouped on task. The random intercept absorbs baseline differences in token cost across tasks, so the fixed effects capture within-task generational differences in tokens-to-solve.

\paragraph{Reliability analysis.}
Reliability asks whether later generations solve already-unlocked tasks more consistently across repeated attempts. We therefore analyse the same unlocked (model, task) cells, with solve rate $s_{m,t}$ as the outcome. For each benchmark, we regress per-task solve rate on model generation, leave-one-model-out task difficulty, and their interaction by OLS with cluster-robust standard errors at the task level. In Wilkinson notation,
\[
\texttt{solve\_rate} \sim \texttt{generation} * \texttt{task\_pass\_rate},
\]
or equivalently
\begin{equation*}
s_{m,t} = \alpha + \beta_{\mathrm{gen}} g_m + \beta_{\mathrm{pass}} \pi_{-m,t} + \beta_{\mathrm{int}}\, g_m\pi_{-m,t} + \varepsilon_{m,t}.
\end{equation*}
The leave-one-model-out difficulty predictor $\pi_{-m,t}$ ensures that the focal model's own solve rate does not appear on both sides of the regression.

\paragraph{Coefficient interpretation.}
Because higher $\pi_t$ and $\pi_{-m,t}$ denote easier tasks, the interaction term $\beta_{\mathrm{int}}$ is interpreted as follows. In the efficiency regression, $\beta_{\mathrm{int}}<0$ indicates that newer models' token savings are larger on harder tasks, while $\beta_{\mathrm{int}}>0$ indicates larger savings on easier tasks. In the reliability regression, $\beta_{\mathrm{int}}<0$ indicates that solve-rate gains are concentrated on harder tasks, while $\beta_{\mathrm{int}}>0$ indicates gains concentrated on easier tasks. The reach regression follows the same convention: $\beta_{\mathrm{int}}<0$ indicates that the generational increase in unlock probability is larger on harder tasks, while $\beta_{\mathrm{int}}>0$ indicates a larger increase on easier tasks.

\subsubsection{Balanced-panel sensitivity for the efficiency analysis}
\label{app:efficiency-balanced-panel}

The efficiency analysis in the main text uses all unlocked \emph{(model, task)} cells. A row enters if that model unlocks that task and has a defined median tokens-to-solve value. This design estimates token use \emph{conditional on reach}, but it does not require that all models solve the same tasks. A potential concern is therefore compositional. Newer models unlock harder tasks, and those tasks may intrinsically require more tokens, which could attenuate measured efficiency gains relative to an analysis restricted to a common solved-task set.

To assess this, we repeated the efficiency regression on a balanced panel within each benchmark, restricting to tasks unlocked by all models evaluated on that benchmark. Coverage in the balanced panel, reported relative to each benchmark's full task set, varied substantially across benchmarks: 27/71 tasks for Cyber CTFs (38.0\%), 50/86 for TerminalBench (58.1\%), 71/100 for SWE-Bench Pro (71.0\%), 22/100 for HealthBench (22.0\%), and 39/100 for HLE (39.0\%). FrontierMath has only 4/12 tasks unlocked by all models, below our five-task minimum for a refit, so it admits no balanced-panel estimate. On Cyber CTFs, every task unlocked by all models is solved at the maximum cross-model rate, so task difficulty is constant on the balanced set and the refit identifies only the generation main effect ($\beta_{\mathrm{gen}}$), not the generation-by-difficulty interaction.

Across benchmarks, the main effect of model generation on efficiency ($\beta_{\mathrm{gen}}$) was qualitatively robust on Cyber CTFs, TerminalBench, SWE-Bench Pro, and HealthBench: later generations continued to use fewer tokens on the shared solved-task set, with effect sizes attenuated, strengthened, or essentially unchanged across benchmarks but never reversed (\Cref{tab:efficiency-balanced-panel}); HLE remained non-significant in both panels. The generation-by-difficulty interaction term ($\beta_{\mathrm{int}}$) was substantially less stable, changing sign on TerminalBench, strengthening on HLE, and moving toward zero on SWE-Bench Pro (on Cyber CTFs the balanced task set has no difficulty variation, so the interaction is not identified), so fine-grained conclusions about whether efficiency gains concentrate on easier or harder tasks are sensitive to panel composition.

These refits therefore support the benchmark-level conclusion that later generations often use fewer tokens on tasks they can solve, while indicating that the main-text efficiency estimates should be read as conditional-on-reach effects rather than pure within-task efficiency improvements on a fixed common task set. The interaction estimates by task difficulty appear less robust and are not a major basis for our conclusions.

\begin{table}[!h]
\centering
\begin{tabular}{lrrrrr}
\toprule
Benchmark & Tasks retained & $\beta_{\mathrm{gen}}$ main & $\beta_{\mathrm{gen}}$ balanced & $\beta_{\mathrm{int}}$ main & $\beta_{\mathrm{int}}$ balanced \\
\midrule
Cyber CTFs    & 27 / 71 & $-0.599^{***}$ & $-0.396^{***}$ & $+0.766^{*}$ & ---           \\
TerminalBench & 50 / 86  & $-0.174^{***}$ & $-0.215^{***}$ & $-0.352$      & $+0.060$      \\
SWE-Bench Pro & 71 / 100  & $-0.092^{**}$  & $-0.106^{***}$ & $-0.432$      & $-0.068$      \\
HLE           & 39 / 100  & $+0.030$       & $-0.030$       & $-0.358$      & $-0.814^{*}$ \\
HealthBench   & 22 / 100  & $-0.294^{***}$ & $-0.204^{**}$  & $+0.203$      & $+0.183$      \\
FrontierMath  & 4 / 12   & $-0.165$       & ---            & $+1.323$      & ---           \\
\bottomrule
\end{tabular}
\caption{Balanced-panel sensitivity check for the efficiency analysis. Main-panel regressions use all unlocked \emph{(model, task)} cells. Balanced-panel regressions restrict to tasks unlocked by all models within a benchmark; tasks retained are reported relative to each benchmark's full task set. Estimates are coefficients on model generation ($\beta_{\mathrm{gen}}$) and the generation-by-difficulty interaction ($\beta_{\mathrm{int}}$) from the efficiency regression described in the main text.}
\label{tab:efficiency-balanced-panel}
\end{table}

\subsection{Serial scaling: supplementary methods and results}
\label{app:serial-scaling-methods}

\subsubsection{Cumulative success over submissions}

For a trajectory with submissions indexed by $k = 1,2,\dots$, let $C_i(k)$ indicate whether at least one of the first $k$ submissions is correct:
\begin{equation}
C_i(k) := \mathbf{1}\!\left(\exists j \le k : \text{submission } j \text{ is correct}\right).
\end{equation}
The cumulative success curve for a set of trajectories is the mean of $C_i(k)$ over the trajectories in that set. In \Cref{fig:serial-cumulative}A, these curves are first averaged within model and then pooled across models with equal weight within each benchmark--condition cell.

We summarise repeated-submission gains using three descriptive quantities:
\begin{itemize}
    \item \textbf{Iteration gain:} cumulative success at the highest observed submission index minus cumulative success at the first submission
    \item \textbf{Uplift multiplier:} cumulative success at the highest observed submission index divided by cumulative success at the first submission
    \item \textbf{$k_{90}$:} the smallest submission index at which the cumulative success curve reaches 90\% of its total iteration gain
\end{itemize}

For plotting, the submission axis in \Cref{fig:serial-cumulative}A is truncated at the smallest submission index for which the per-step gain falls below 0.1 percentage points, computed separately for each condition within benchmark and then taking the maximum across conditions, with a two-step tail added for visual context.

\subsubsection{Iteration uplift and model generation}
\label{app:uplift-analysis}

For each benchmark--model--condition cell, we define \emph{iteration uplift} as
\begin{equation}
\Delta = S_{\max} - S_{1},
\end{equation}
where $S_{1}$ is cumulative success at the first submission and $S_{\max}$ is the highest cumulative success level reached over the observed submission range. We then compute Kendall rank correlations (Kendall's $\tau$) between $\Delta$ and model release date separately for each benchmark and condition. The per-condition trend lines in \Cref{fig:serial-cumulative}B are Theil--Sen (median-slope) fits, whose slope sign agrees with $\tau$ by construction. These correlations are descriptive only, since each benchmark--condition cell contains six models.

\begin{table}[!h]
\centering
\begin{tabular}{lcc}
\toprule
Benchmark & Self-guided $\tau$ & Oracle-guided $\tau$ \\
\midrule
TerminalBench & $-0.276$ & $-0.200$ \\
SWE-Bench Pro & $+0.067$ & $+0.067$ \\
FrontierMath  & $-0.276$ & $+0.200$ \\
HLE           & $-0.467$ & $-0.067$ \\
HealthBench   & $+0.333$ & $+0.200$ \\
\bottomrule
\end{tabular}
\caption{Kendall's $\tau$ rank correlations between model release date and iteration uplift, computed separately for each benchmark and feedback condition. Iteration uplift is the increase in cumulative success from the first submission to the highest observed cumulative level. These correlations are descriptive only, since each benchmark--condition cell contains six models.}
\label{tab:serial-uplift-correlations}
\end{table}

\subsubsection{Regression models for eventual success and submissions-to-success}
\label{app:serial-regressions}

For each benchmark separately, we fit regressions with task fixed effects and cluster-robust standard errors at the task level. The feedback indicator $O_i$ takes value 1 under oracle score feedback and 0 under no feedback. Model generation $G_i$ is encoded as a normalised release-rank variable within benchmark, increasing from the oldest to the newest model.

\paragraph{Model A: eventual success.}
We model whether trajectory $i$ on task $s(i)$ ever produces a correct submission:
\begin{equation}
\Pr(Y_i = 1)
=
\mathrm{logit}^{-1}\!\left(
\alpha_{s(i)} + \beta_O O_i + \beta_G G_i + \beta_{G\times O}(G_i O_i)
\right),
\end{equation}
where $Y_i$ is an indicator for eventual success and $\alpha_{s(i)}$ is a task fixed effect.

\begin{table*}[!h]
\centering
\begin{tabular}{lccc}
\toprule
Benchmark & $\beta_O$ [95\% CI] & $\beta_G$ [95\% CI] & $\beta_{G\times O}$ [95\% CI] \\
\midrule
FrontierMath  & $-0.034$ [$-0.353, +0.284$] & $+2.851$ [$+1.862, +3.840$] & $+0.521$ [$-0.047, +1.090$] \\
SWE-Bench Pro & $+0.247$ [$+0.128, +0.365$] & $+0.752$ [$+0.481, +1.022$] & $-0.175$ [$-0.324, -0.027$] \\
HealthBench   & $+0.062$ [$-0.088, +0.212$] & $+0.355$ [$+0.144, +0.565$] & $+0.048$ [$-0.176, +0.273$] \\
HLE           & $+0.785$ [$+0.578, +0.993$] & $+1.065$ [$+0.729, +1.401$] & $-0.156$ [$-0.427, +0.115$] \\
TerminalBench & $+0.016$ [$-0.148, +0.179$] & $+1.094$ [$+0.784, +1.405$] & $+0.210$ [$-0.048, +0.468$] \\
\bottomrule
\end{tabular}
\caption{Model A: fixed-effects logistic regressions of eventual success. The response is whether a trajectory ever produces a correct submission. Predictors are oracle feedback ($O$), normalised model generation ($G$), and their interaction.}
\label{tab:serial-modelA}
\end{table*}

\paragraph{Model B: submissions to first correct answer.}
Conditioning on trajectories that eventually succeed, we model the submission index of the first correct answer:
\begin{equation}
\mathbb{E}[K_i]
=
g^{-1}\!\left(
\alpha_{s(i)} + \beta_O O_i + \beta_G G_i + \beta_{G\times O}(G_i O_i)
\right),
\end{equation}
where $K_i$ is the submission index of the first correct answer, $\alpha_{s(i)}$ is a task fixed effect, and $g^{-1}$ is the inverse link function. We use a negative-binomial model for HLE and Poisson models for the remaining benchmarks, based on Pearson dispersion.

Before fitting Model B, we remove successful trajectories whose total number of submissions is more than 3 standard deviations from the within-benchmark mean, excluding 146 of 13{,}366 successful trajectories and leaving 13{,}220 for analysis.

\begin{table*}[!h]
\centering
\begin{tabular}{lcccc}
\toprule
Benchmark & Family & $\beta_O$ [95\% CI] & $\beta_G$ [95\% CI] & $\beta_{G\times O}$ [95\% CI] \\
\midrule
FrontierMath  & Poisson      & $-0.049$ [$-0.215, +0.117$] & $-0.162$ [$-0.314, -0.010$] & $+0.108$ [$-0.088, +0.303$] \\
SWE-Bench Pro & Poisson      & $+0.049$ [$+0.009, +0.089$] & $-0.005$ [$-0.062, +0.051$] & $-0.037$ [$-0.086, +0.013$] \\
HealthBench   & Poisson      & $+0.041$ [$-0.028, +0.111$] & $-0.008$ [$-0.101, +0.084$] & $-0.001$ [$-0.101, +0.099$] \\
HLE           & Neg.-bin.    & $+0.325$ [$+0.108, +0.542$] & $-0.594$ [$-0.935, -0.254$] & $+0.037$ [$-0.268, +0.341$] \\
TerminalBench & Poisson      & $+0.065$ [$-0.043, +0.173$] & $-0.148$ [$-0.297, +0.001$] & $-0.010$ [$-0.162, +0.141$] \\
\bottomrule
\end{tabular}
\caption{Model B: regressions of submission index at first correct answer, conditional on eventual success. Negative coefficients indicate fewer submissions to the first correct answer.}
\label{tab:serial-modelB}
\end{table*}

\subsection{Parallel scaling: supplementary methods and validation}

\subsubsection{Pass@$k$ estimator and pooling}
\label{app:parallel-scaling-methods}

For total budget $B$, we estimate pass@$k$ by splitting $B$ across $k$ independent trajectories of $B/k$ tokens each and counting a task as solved if any trajectory's first correct submission falls within budget. pass@$k$ is computed task-wise using the unbiased estimator of \citet{chen2021codex}. For the main analysis, we pool self-guided and oracle trajectories because the first submission is pre-feedback, yielding up to ten trajectories per (benchmark, model, task). Figure~\Cref{fig:parallel-scaling} reports benchmark-level means across models with equal weighting. We use pass@$k$ rather than consensus-based aggregation because our aim is to measure the value of independent restarts under matched total compute, and several benchmarks involve stateful trajectories or structured outputs for which a benchmark-agnostic consensus rule is not well defined.

\subsubsection{Judge reliability validation for HLE and HealthBench}
\label{app:scorer-reliability}

Two of our five primary benchmarks, HLE and HealthBench, are scored with an LLM judge, GPT-4o-mini at temperature 0, rather than a deterministic verifier. A potential confound for the parallel-scaling results in \Cref{sec:parallel-scaling} is that, if the judge has a per-submission false-positive rate $p_\mathrm{FP}$, then $N$ independent parallel attempts give a worst-case spurious inflation of $1 - (1 - p_\mathrm{FP})^N$, which grows with $N$. To bound this, we re-scored a stratified subsample of trajectories with two stronger judges, \texttt{openai/gpt-5.4} and \texttt{anthropic/claude-opus-4-6}, using the exact in-experiment grader template for each benchmark, \texttt{inspect\_evals.hle.judge.JUDGE\_PROMPT} and \texttt{inspect\_evals.healthbench.scorer.GRADER\_TEMPLATE}. The subsample is 6 models $\times$ 2 conditions $\times$ 20 sample IDs $\times$ up to 3 trajectories per bin: 228 trajectories (2{,}374 submissions) on HLE and 175 trajectories (537 submissions) on HealthBench. Re-running the pipeline with GPT-4o-mini itself as the re-scoring judge reproduces the in-experiment scores to within expected non-determinism, 99.5\% agreement on HLE and Pearson $r = 0.87$ on HealthBench. Any disagreement reported below is therefore attributable to judge choice rather than pipeline drift.

\paragraph{HLE: in-experiment judge is conservative; confound is bounded below the effect.}
The two strong judges agree on 2{,}370 of 2{,}373 paired HLE submissions (99.9\%), and we treat their consensus as a high-confidence reference label. Against this consensus, GPT-4o-mini produces 1 false positive in 2{,}186 actually incorrect submissions, a false-positive rate of $0.05\%$ with 95\% Wilson CI $[0.01\%, 0.26\%]$, and misses 22 of 184 correct submissions, a false-negative rate of $12.0\%$. With $p_\mathrm{FP} = 5\times 10^{-4}$, the worst-case independent-error parallel-scaling inflation is $0.5\%$ at $N = 10$ and $4.9\%$ at $N = 100$. This is an order of magnitude below the observed parallel gains on HLE (\Cref{tab:parallel-gains}). Errors are also strongly clustered at the sample level, with index of dispersion 7.81, and 95.7\% of disagreements come from 3 of the 20 sample IDs. This further reduces the effective number of independent dice rolls and tightens the bound.

\paragraph{HealthBench: judge noise is large but unbiased.}
HealthBench is continuously scored and noisier. Pairwise Pearson $r$ between strong judges is $0.78$, and mean per-judge scores differ by up to $0.15$. Restricting to the 271 submissions on which the two strong judges agree within $0.15$, the signed error $\Delta = (\text{GPT-4o-mini}) - (\text{strong-judge consensus})$ has mean $-0.015$ and standard deviation $0.239$. Tail inflations, $\Delta > 0.20$, occur on 17.0\% of submissions and are statistically indistinguishable from tail deflations, $\Delta < -0.20$, which occur on 18.1\%. This is consistent with symmetric noise rather than directional bias. Because the parallel-scaling statistic averages over many submissions, the symmetric per-submission noise cancels in expectation, and per-sample clustering (index of dispersion 3.14) further reduces the effective number of independent attempts. The HLE result above, with a per-submission false-positive rate too small to drive meaningful inflation, provides the strongest non-confound test; taken together, the parallel-scaling gains on HLE and HealthBench cannot be explained by judge unreliability.

\end{document}